\documentclass{article}

    \PassOptionsToPackage{numbers, compress}{natbib}
    \usepackage[preprint]{neurips_2023}

\usepackage[utf8]{inputenc} % allow utf-8 input
\usepackage[T1]{fontenc}    % use 8-bit T1 fonts
\usepackage{hyperref}       % hyperlinks
\usepackage{url}            % simple URL typesetting
\usepackage{booktabs}       % professional-quality tables
\usepackage{amsfonts}       % blackboard math symbols
\usepackage{nicefrac}       % compact symbols for 1/2, etc.
\usepackage{microtype}      % microtypography
\usepackage{graphicx}
\usepackage{booktabs} % for professional tables
\usepackage{amsmath}
\usepackage{subcaption}
\usepackage{wrapfig}
\usepackage{array}

\usepackage[dvipsnames]{xcolor}

\newif\ifcomments
\commentstrue
\ifcomments
  \newcommand{\colornote}[3]{{\color{#1}\bf{#2: #3}\normalfont}}
\else
  \newcommand{\colornote}[3]{}
\fi

\newcommand{\generator}{\mathcal{G}}

\renewcommand{\paragraph}[1]{\textbf{#1}\hspace{0.8em}}
\newcolumntype{L}[1]{>{\raggedright\let\newline\\\arraybackslash\hspace{0pt}}m{#1}}
\newcolumntype{C}[1]{>{\centering\let\newline\\\arraybackslash\hspace{0pt}}m{#1}}
\newcolumntype{R}[1]{>{\raggedleft\let\newline\\\arraybackslash\hspace{0pt}}m{#1}}

\title{Training on Thin Air: Improve Image Classification with Generated Data}

\author{%
  Yongchao Zhou \\
  University of Toronto\\
  \texttt{yczhou@cs.toronto.edu} \\
  \And
  Hshmat Sahak \\
  University of Toronto\\
  \texttt{hshmat.sahak@mail.utoronto.ca} \\
  \And
  Jimmy Ba \\
  University of Toronto\\
  \texttt{jba@cs.toronto.edu} \\
}

\begin{document}

\maketitle

% Keywords. diffusion model, data augmentation, synthetic data, image classification, generative model, model inversion, few-shot learning, medical imaging

% TL;DR. We introduce Diffusion Inversion, a simple yet effective approach that utilizes pre-trained generative models to produce high-quality training data for image classification tasks, surpassing conventional data augmentation methods and enhancing sample efficiency.
\begin{abstract}
Acquiring high-quality data for training discriminative models is a crucial yet challenging aspect of building effective predictive systems. In this paper, we present Diffusion Inversion, a simple yet effective method that leverages the pre-trained generative model, Stable Diffusion, to generate diverse, high-quality training data for image classification. Our approach captures the original data distribution and ensures data coverage by inverting images to the latent space of Stable Diffusion, and generates diverse novel training images by conditioning the generative model on noisy versions of these vectors. We identify three key components that allow our generated images to successfully supplant the original dataset, leading to a 2-3x enhancement in sample complexity and a 6.5x decrease in sampling time. Moreover, our approach consistently outperforms generic prompt-based steering methods and KNN retrieval baseline across a wide range of datasets. Additionally, we demonstrate the compatibility of our approach with widely-used data augmentation techniques, as well as the reliability of the generated data in supporting various neural architectures and enhancing few-shot learning~\footnote{Please visit our project page at \url{https://sites.google.com/view/diffusion-inversion}}.
\end{abstract}

\section{Introduction}
Collecting data from the real world can be complex, costly, and time-consuming. Traditional machine learning datasets are often not curated, and noisy, or hand-curated but lacking size. Consequently, though obtaining high-quality data is critical, it remains a challenging aspect of developing effective predictive systems. Recently, large-scale machine learning models such as GPT-3~\cite{brown2020language}, DALL-E~\cite{ramesh2022hierarchical}, Imagen~\cite{saharia2022photorealistic}, and Stable Diffusion~\cite{rombach2022high}, which are trained on vast amounts of noisy internet data, have emerged as successful "foundation models"~\cite{bommasani2021opportunities} demonstrating strong generative capabilities such as co-authoring code, creating art, and writing text. Given their extensive world knowledge, a natural question arises: can large-scale pre-trained generative models help generate high-quality training data for discriminative models? 
% \textcolor{red}{that achieves higher classification accuracy score than the original dataset?}

In computer vision, generative models have long been considered for data augmentation. Previous work has explored using VAEs~\citep{shrivastava2017learning, viazovetskyi2020stylegan2}, GANs~\citep{antoniou2017data,chai2021ensembling}, and Diffusion Models~\citep{antoniou2017data} to enhance model performance in data-scarce settings, such as zero-shot or few-shot learning~\citep{antoniou2017data,he2022synthetic}, or to enhance robustness against adversarial attacks or natural distribution shifts~\citep{bansal2023leaving, gowal2021improving}. However, due to the limited diversity in samples generated by previous approaches, it has been widely believed and empirically observed that these samples cannot be utilized to train classifiers with higher absolute accuracy compared to those trained on the original datasets~\citep{gowal2021improving, ravuri2019classification, ravuri2019seeing,  zhao2022synthesizing}. Nevertheless, the issue of generator quality may no longer be a hindrance, as state-of-the-art diffusion-based text-to-image models demonstrate remarkable capabilities in synthesizing diverse images with high visual fidelity. 

A natural approach to utilizing these models to augment the original dataset involves incorporating human language intervention. By applying prompt engineering, domain expert knowledge about the target domain can be distilled into several sets of prompts. When combined with language enhancement techniques~\citep{he2022synthetic, yuan2022not}, a diverse array of high-fidelity images can be generated. However, despite their diversity, a prompt-based generation often produces off-topic and irrelevant images from the target domain, resulting in low-quality datasets~\citep{bansal2023leaving}. To mitigate the issue of low-quality images generated by prompts, CLIP filtering~\citep{he2022synthetic} has been introduced, enabling a more favorable balance between prompt diversity and quality. Nonetheless, the generation process continues to disregard the distribution of the training dataset, leading to the creation of distributionally dissimilar images compared to the original data, resulting in a substantial gap between real and synthetic datasets~\citep{borji2022good}. Furthermore, although in-distribution examples can be generated infinitely, the generated data must still provide sufficient coverage of the original dataset to yield optimal performance.

\begin{figure}[t]
\centering
\begin{subfigure}[b]{0.725\textwidth}
\centering
\includegraphics[width=1.0\linewidth]{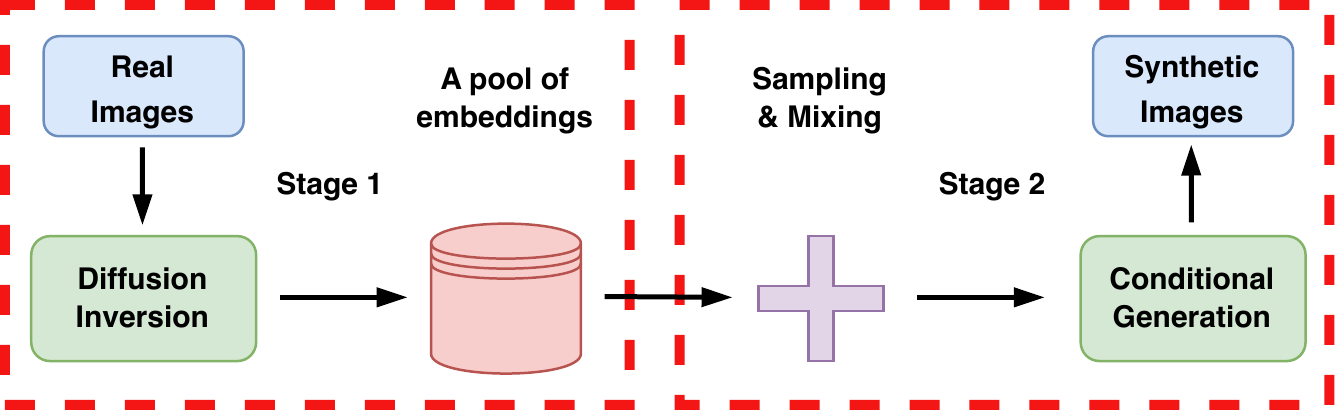}
\end{subfigure}
\hfill
\begin{subfigure}[b]{0.255\textwidth}
\centering
\includegraphics[width=1.0\linewidth]{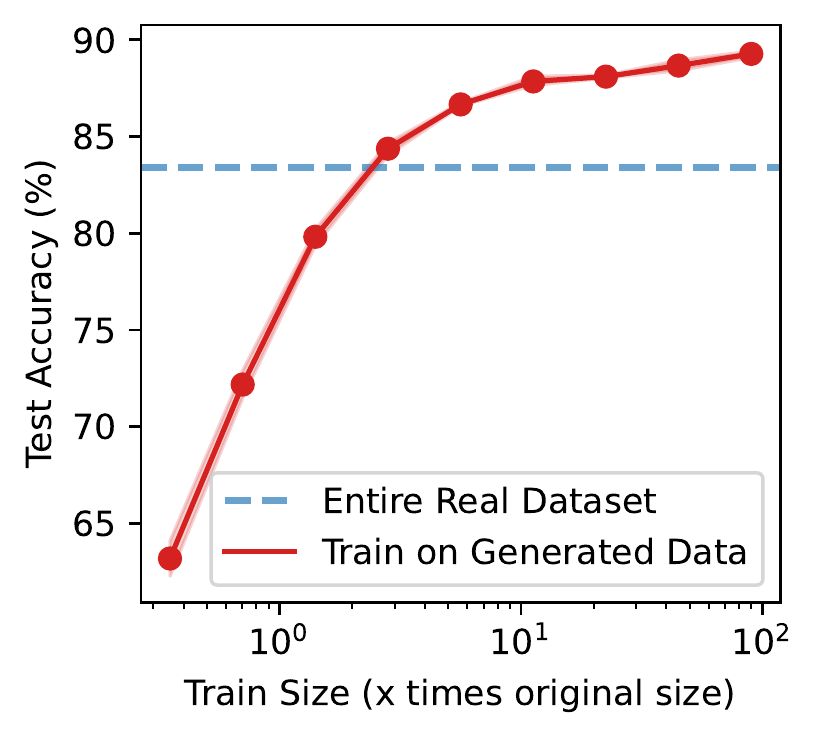}
% \caption{Scale the Dataset Size}
\end{subfigure}
\caption{\footnotesize{(Left) Our two-stage approach utilizes Stable Diffusion's generalizable knowledge for targeted classification tasks by transforming real images into latent space and generating novel variants through inverse diffusion with perturbed embeddings. (Right) The test accuracy of ResNet18 increases as more generated data is incorporated, eventually exceeding the performance of the model trained on the entire real dataset.}}
\vspace{-0.15in}
\label{fig:workflow}
\end{figure}

To address these challenges and close the performance gap between generated and real data, we present Diffusion Inversion, a simple yet effective method that leverages the general-purpose pre-trained image generator, Stable Diffusion~\citep{rombach2022high}. To capture the original data distribution and ensure data coverage, we first obtain a set of embedding vectors by inverting each training image to the output space of the text encoder. Next, we condition Stable Diffusion on a noisy version of these vectors, enabling sampling of a diverse array of novel training images extending beyond the initial dataset. As a result, the final generated images retain semantic meaning while incorporating variability stemming from the rich knowledge embedded within the pre-trained image generator (see examples in Figure~\ref{fig:syn_data_vis} and~\ref{fig:ab_latent_vis}). Furthermore, we enhance sampling efficiency by learning condition vectors to generate low-resolution images directly rather than producing them at high resolution and subsequently downsampling. This strategy increases the generation speed of the diffusion model by 6.5 times, rendering it more suitable as a data augmentation tool. To assess our method, we compare it against generic prompt-based steering methods, widely-used data augmentation techniques, and original real data across various datasets. Our primary contributions include:
\begin{itemize}
    \item We propose Diffusion Inversion, a simple yet effective method that utilizes pre-trained generative models to assist with discriminative learning, bridging the gap between real and synthetic data. Our method offers 2-3x sample complexity improvements and 6.5x reduction in sampling time.
    \item We pinpoint three vital components that allow models trained on generated data to surpass those trained on real data: 1) a high-quality generative model, 2) a sufficiently large dataset size, and 3) a steering method that considers distribution shift and data coverage.
    \item Our method outperforms generic prompt-based steering methods and widely-used data augmentation techniques, especially in the realm of specialized datasets such as medical imaging, exhibiting significant data distribution shifts. Additionally, our generated data can enhance various neural architectures and boost few-shot learning performance.
\end{itemize}

\section{Related Work}

\paragraph{Utilizing Generative Models for Image Data Augmentation}
Generative models, such as VAEs~\citep{kingma2013auto}, GANs~\citep{goodfellow2020generative}, and Diffusion Models~\cite{dhariwal2021diffusion}, have exhibited exceptional capabilities in synthesizing realistic images. Due to their potential to generate an infinite amount of high-quality data, numerous researchers have investigated their application as data augmentation techniques. For example, several works~\citep{shrivastava2017learning,viazovetskyi2020stylegan2,zhu2017data}~formulate data augmentation as an image translation task, training an autoencoder-style network to produce multiple variations of input images for downstream prediction models. Some studies have concentrated on data augmentation using GANs, either training them from scratch for few-shot learning~\citep{antoniou2017data} or utilizing pre-trained GANs for self-supervised learning~\cite{chai2021ensembling}. Despite their effectiveness in various domains, research has shown that training off-the-shelf convolutional networks, such as ResNet50~\citep{he2016deep}, on BigGAN~\citep{brock2018large} synthesized images yields inferior results compared to training them on original real training images due to the lack of diversity and the potential domain gap between generated samples and real images~\citep{bansal2023leaving,gowal2021improving,ravuri2019classification,zhao2022synthesizing}. 

\paragraph{Augmenting Image Data using Text-to-Image Models}
Recently, there has been a growing interest in leveraging the power of internet-scale pre-trained diffusion-based models~\citep{rombach2022high, nichol2021glide} for data generation. \citet{he2022synthetic} demonstrates that synthetic data from GLIDE~\citep{nichol2021glide} can enhance classification models in data-scarce settings or pre-training. Meanwhile, several works \citep{bansal2023leaving,yuan2022not,sariyildiz2023fake} illustrate that Stable Diffusion~\citep{rombach2022high} can serve as a data augmentation tool to learn generalizable features and improve the robustness of image classifiers under natural distribution shifts. However, the effectiveness of these approaches largely relies on the quality and diversity of language prompts, necessitating extensive manual prompt engineering. Furthermore, the domain gap between synthetic and real data in downstream tasks may continue to hinder the improvement of synthetic data's effectiveness in classifier learning \citep{he2022synthetic,bansal2023leaving}. To enhance the alignment of the text-to-image model with the downstream dataset, \citet{azizi2023synthetic} suggests fine-tuning the model weights and sampling parameters while retaining the text prompts as concise one or two-word class names from \cite{radford2021learning}. Nonetheless, generative diversity and data coverage may still present obstacles, resulting in the generation of data that is inferior to real data. 
In contrast, our approach addresses these issues by directly learning the conditioning vector for each target image and producing new variants by conditioning on noisy versions of these vectors. This method eliminates the need for human prompt engineering, guarantees data coverage, and promotes diversity.

\paragraph{Inversion Techniques in Generative Models}
Inverting generative models plays a crucial role in image editing and manipulation tasks~\citep{zhu2016generative, xia2022gan, creswell2018inverting}. For diffusion models, inversion can be accomplished by adding noise to an image and subsequently denoising it through the network. However, this may result in significant content alterations due to the asymmetry between backward and forward diffusion steps. \citet{choi2021ilvr} address inversion by conditioning the denoising process on noisy, low-pass filtered data from the target image. More recently, inverting text-to-image diffusion models in the context of personalized image generation has gained traction. \citet{gal2022image} propose a textual inversion method that learns to represent visual concepts through new pseudo-words in the embedding space of a frozen text-to-image model. In contrast, \citet{ruiz2022dreambooth} fine-tune the entire network on 3-5 images, which may be susceptible to overfitting. Custom Diffusion~\citep{kumari2022multiconceptco} mitigates overfitting by fine-tuning only a small subset of model parameters, resulting in improved performance with reduced training time. These works employ inversion as a tool for image editing and have only assessed qualitative human preferences. In contrast, our work seeks to explore how generated images can enhance downstream image classification tasks and proposes using diffusion inversion to address the distribution shift and data coverage problem in synthetic dataset generation.
% Particularly, given an input image, inversion algorithms~\citep{creswell2018inverting, xia2022gan} strive to determine the latent representation within the generator that reconstructs the original input. 
\section{Method}\label{sec:method}
Stable Diffusion~\citep{rombach2022high}, a model trained on billions of image-text pairs, boasts a wealth of generalizable knowledge. To harness this knowledge for specific classification tasks, we propose a two-stage method that guides a pre-trained generator, $\generator$, towards the target domain dataset. In the first stage, we map each image to the model's latent space, generating a dataset of latent embedding vectors. Then, we produce novel image variants by running the inverse diffusion process conditioned on perturbed versions of these vectors. We illustrate our approach in Figure~\ref{fig:workflow} (Left).
% We outline the two stages below and conclude with a discussion on the method's runtime.

\subsection{Stage 1 - Embedding Learning}
\paragraph{Stable Diffusion}
Stable Diffusion is a type of Latent Diffusion Model (LDM), which is a class of Denoising Diffusion Probabilistic Models. LDMs operate in an autoencoder's latent space and have two main components. First, an autoencoder is pre-trained on a large image dataset to minimize reconstruction loss, using regularization from either KL-divergence loss or vector quantization~\cite{van2017neural, agustsson2017soft}. This allows the encoder $\mathcal{E}$ to map images $x \in \mathcal{D}_x$ to a spatial latent code $z=\mathcal{E}(x)$, while the decoder $D$ converts these latents back into images, such that $D(\mathcal{E}(x)) \approx x$. Next, a diffusion model is trained to minimize the denoising objective in the derived latent space, incorporating optional conditional information from class labels, segmentation masks, or text tokens.
\begin{equation*}
    L_{L D M}:=\mathbb{E}_{z \sim \mathcal{E}(x), y, \epsilon \sim \mathcal{N}(0,1), t}\left[\left\|\epsilon-\epsilon_\theta\left(z_t, t, c_\theta(y)\right)\right\|_2^2\right],
\end{equation*}
where $t$ represents the time step, $z_t$ denotes the latent noise at time $t$, $\epsilon$ is the unscaled noise sample, $\epsilon_\theta$ denotes the denoising network, and $c_\theta(y)$ is a model mapping conditioning input $y$ to a vector. During inference, a new image latent $z_0$ is generated by iteratively denoising a random noise vector with a conditioning vector, and the latent code is transformed into an image using the pre-trained decoder $x^{\prime}=D\left(z_0\right)$.

\begin{wrapfigure}{r}{0.45\textwidth}
  \centering
  \vspace{-0.15in}
  \includegraphics[width=1.0\linewidth]{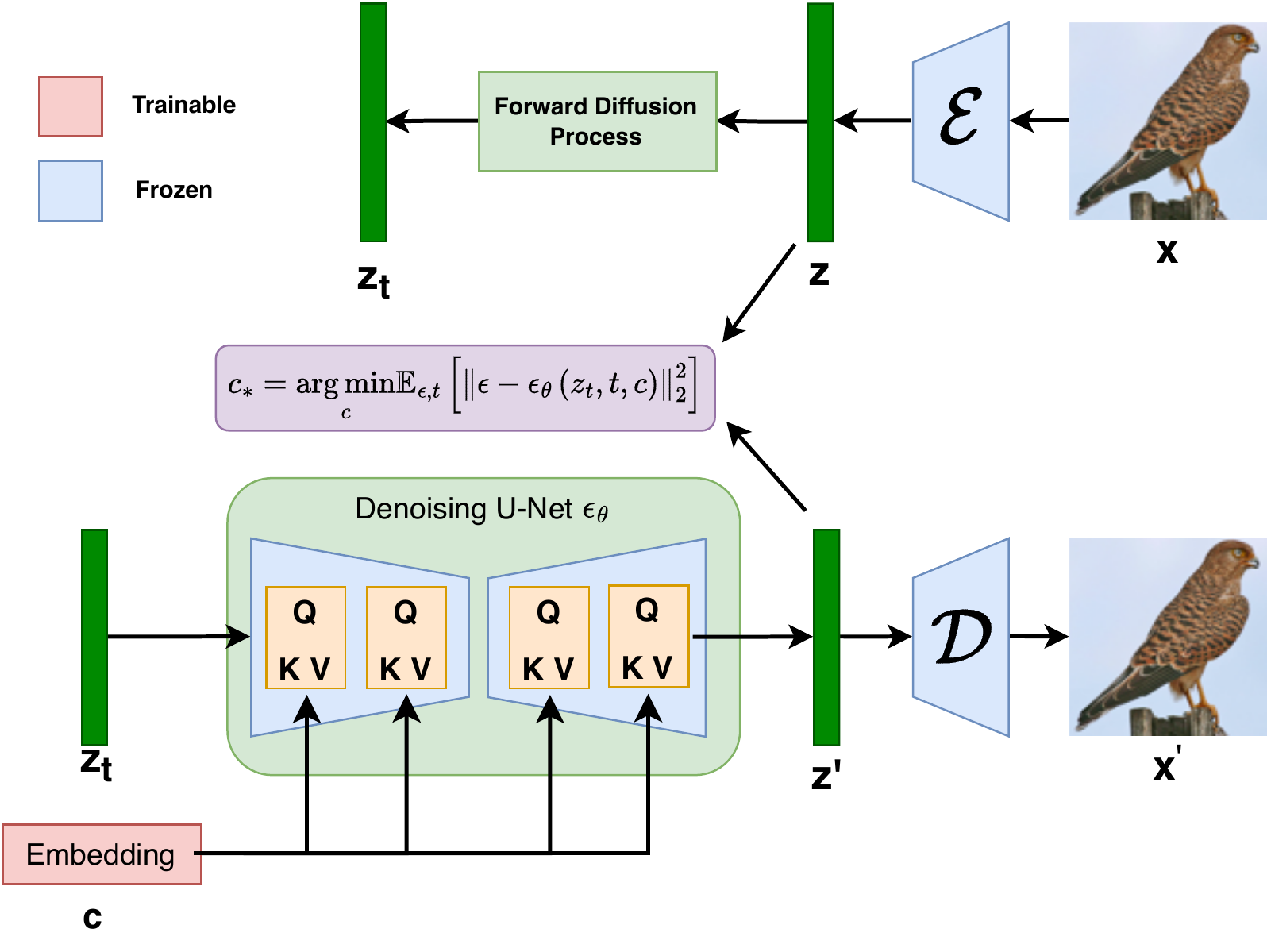}
  \vspace{-0.15in}
  \caption{\footnotesize{Our method optimizes the standard denoising objective to learn a set of embedding vectors while keeping the model parameters fixed.}}
  \label{fig:method_illustration}
  \vspace{-0.05in}
\end{wrapfigure}

\paragraph{Diffusion Inversion}
Prior research has attempted to invert images back to the input tokens of a text encoder $c_\theta$~\cite{gal2022image}. However, this approach is restricted by the expressiveness of the textual modality and constrained to the original output domain of the model. To overcome this limitation, we treat $c_\theta$ as an identity mapping and directly optimize the conditioning vector $c$ for each image latent $z$ in the real dataset by minimizing the LDM loss.
\begin{equation*}\label{method:loss}
    c_*=\underset{c}{\arg \min } \mathbb{E}_{\epsilon \sim \mathcal{N}(0,1), t}\left[\left\|\epsilon-\epsilon_\theta\left(z_t, t, c\right)\right\|_2^2\right],
\end{equation*}
Throughout the optimization process, we maintain the original LDM model's training scheme and keep the denoising model $\epsilon_\theta$ unchanged to optimally maintain pre-training knowledge. Furthermore, we improve sampling efficiency by learning condition vectors tailored to generate target-resolution images, instead of creating high-resolution images and subsequently downsampling, thereby considerably reducing the overall generation time (see Figure~\ref{fig:runtime}).

\subsection{Stage 2 - Sampling}
\paragraph{Classifier-free Guidance}
Classifier-free guidance employs a weight parameter $w \in \mathcal{R}$ to balance sample quality and diversity in class-conditioned diffusion models, commonly used in large-scale models such as Stable Diffusion~\cite{rombach2022high}, GLIDE~\cite{nichol2021glide}, and Imagen~\cite{saharia2022photorealistic}. During sample generation, both the conditional diffusion model $\epsilon_\theta\left(z_t, t, c\right)$ and the unconditional model $\epsilon_\theta\left(z_t, t\right)$ are evaluated. In Stable Diffusion, the conditioning vector is determined by the text encoder's output for an empty string, with the model output at each denoising step given by $\hat \epsilon = (1+w) \epsilon_\theta\left(z_t, t, c\right) - w \epsilon_\theta\left(z_t, t\right)$. However, we find that using an empty string as conditioning input is ineffective for the target domain when the data distribution deviates significantly from the training distribution, particularly when image resolution varies. To address this distribution shift, we instead utilize the average embedding of all learned vectors as the class-conditioning input for unconditional models, with the effectiveness of this design demonstrated in Section~\ref{sec:avgemb}.

\paragraph{Sample Diversity}
Sample diversity is crucial for training downstream classifiers on synthetic data \cite{ravuri2019classification}. To achieve this, we employ various classifier-free guidance strengths and initiate the denoising process with different random noises, generating distinct image variants. We also explore two conditioning vector perturbation methods: Gaussian noise perturbation and latent interpolation. In the Gaussian approach, we add isotropic Gaussian noise to the conditioning vector, yielding a new vector $\hat c = c + \lambda \epsilon$, where $\epsilon \sim \mathcal{N}(0,1)$ and $\lambda$ indicates the perturbation strength. For latent interpolation, we linearly interpolate between two conditioning vectors $c_1$ and $c_2$ to create a new vector: $\hat c = \alpha c_1 + (1-\alpha) c_2$. We assess each component's impact in Section~\ref{sec:gauss_latent}.

\section{Experimental Results}
We first investigate CIFAR10/100~\cite{krizhevsky2009learning}, STL10~\citep{coates2011analysis}, and ImageNette~\citep{imagenette}, focusing on two key factors that enable models trained on generated data to surpass those trained on real data: 1) a high-quality generative model and 2) a sufficiently large dataset size. Next, we compare our approach with generic prompt-based steering techniques~\citep{he2022synthetic} and KNN retrieval from LAION-5B~\citep{schuhmann2022laion}, highlighting the significance of a steering method that addresses distribution shift and data coverage for discriminative downstream tasks. Concurrently, we demonstrate our method's efficacy in few-shot learning scenarios and specialized datasets such as EuroSAT~\citep{helber2019eurosat} and three MedMNISTv2 datasets~\citep{yang2023medmnistv2}. Furthermore, we show that our generated data is compatible with various standard data augmentation strategies and can enhance model performance across numerous popular neural architectures.

\begin{figure*}[t]
% \vspace{-0.12in}
\begin{center}
\begin{subfigure}[b]{0.24\textwidth}
\centering
\includegraphics[width=1.0\linewidth]{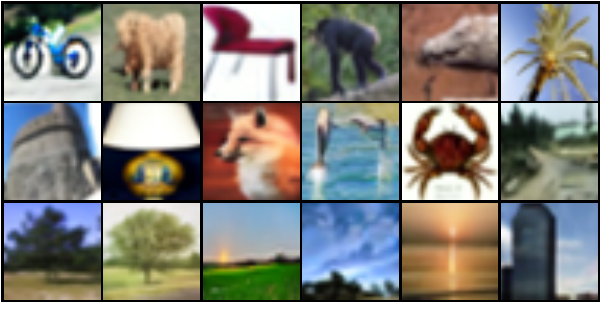}
\caption{CIFAR10}
\end{subfigure}
\begin{subfigure}[b]{0.24\textwidth}
\centering
\includegraphics[width=1.0\linewidth]{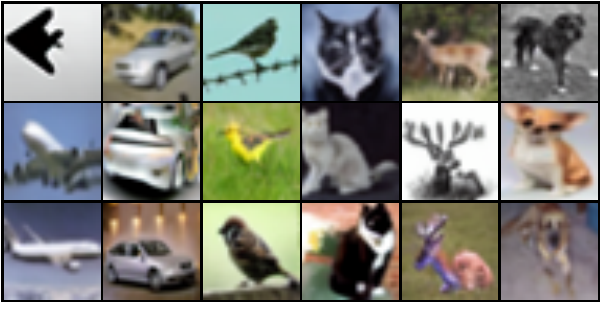}
\caption{CIFAR100}
\end{subfigure}
\begin{subfigure}[b]{0.24\textwidth}
\centering
\includegraphics[width=1.0\linewidth]{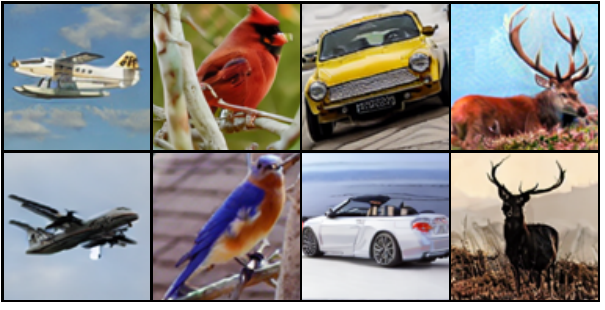}
\caption{STL10}
\end{subfigure}
\begin{subfigure}[b]{0.24\textwidth}
\centering
\includegraphics[width=1.0\linewidth]{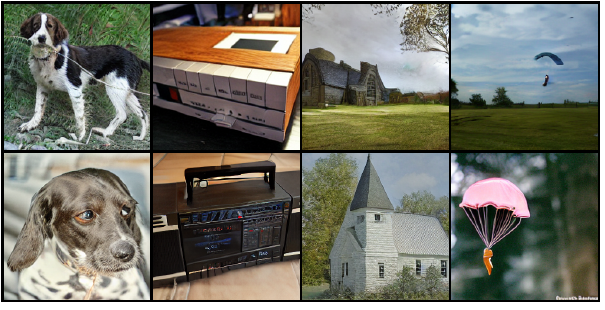}
\caption{ImageNette}
\end{subfigure}
\begin{subfigure}[b]{0.24\textwidth}
\centering
\includegraphics[width=1.0\linewidth]{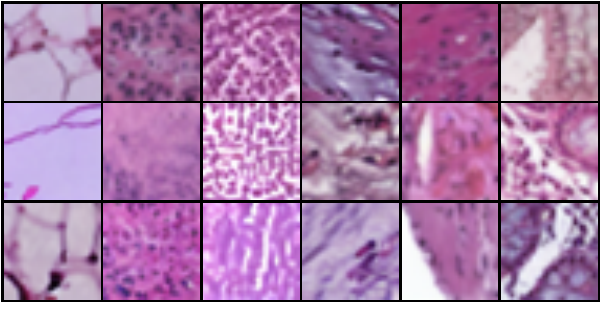}
\caption{PathMNIST}
\end{subfigure}
\begin{subfigure}[b]{0.24\textwidth}
\centering
\includegraphics[width=1.0\linewidth]{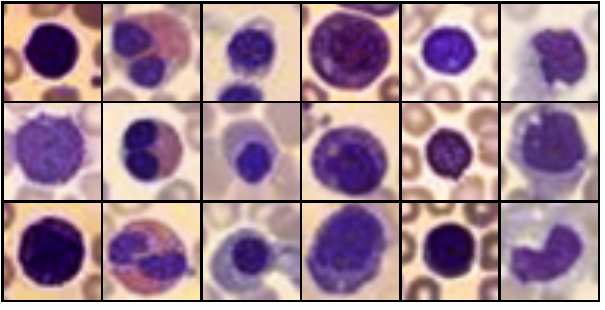}
\caption{BloodMNIST}
\end{subfigure}
\begin{subfigure}[b]{0.24\textwidth}
\centering
\includegraphics[width=1.0\linewidth]{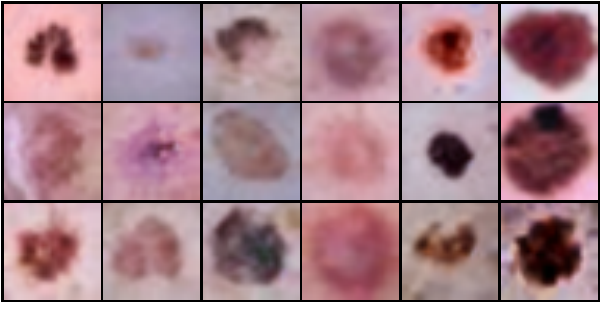}
\caption{DermaMNIST}
\end{subfigure}
\begin{subfigure}[b]{0.24\textwidth}
\centering
\includegraphics[width=1.0\linewidth]{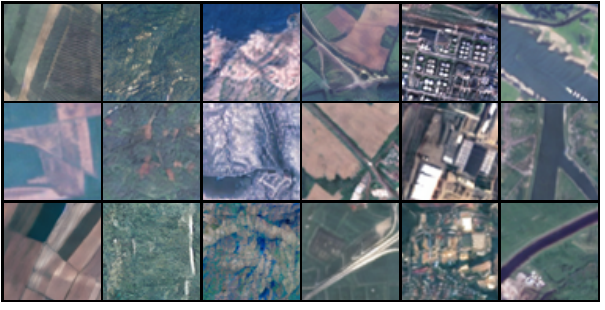}
\caption{EuroSAT}
\end{subfigure}
\end{center}
\vspace{-0.1in}
\caption{\footnotesize{Synthetic images produced by our method: exhibiting diversity, realism, and comprehensive representation of the original dataset, effectively serving as a suitable substitute.}}
\label{fig:syn_data_vis}
\vspace{-0.2in}
\end{figure*}

\begin{wrapfigure}{r}{0.30\textwidth}
  \centering
  \vspace{-0.15in}
  \includegraphics[width=0.9\linewidth]{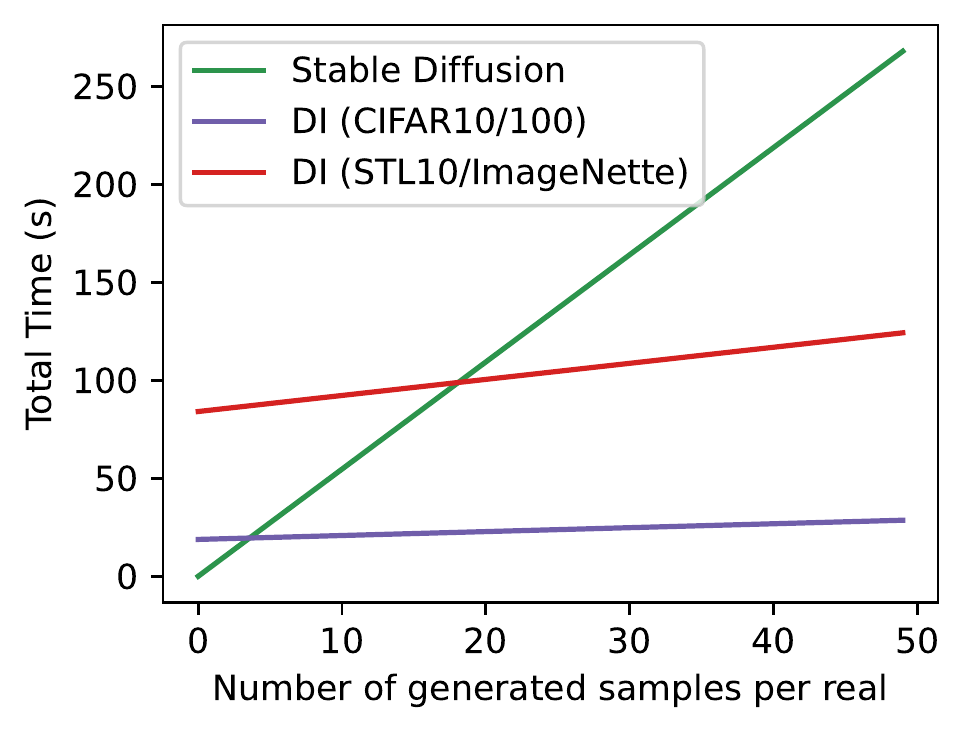}
  \vspace{-0.05in}
  \caption{\footnotesize{Despite the overhead incurred by embedding learning, our method substantially decreases the overall time required to generate numerous images due to improved sampling.}}
  \label{fig:runtime}
  \vspace{-0.05in}
\end{wrapfigure}

We employ the Stable Diffusion model with a default resolution of 512x512 \footnote{We use the checkpoint "CompVis/stable-diffusion-v1-4" from Hugging Face. \url{https://huggingface.co/CompVis/stable-diffusion-v1-4}}. To optimize learning and sampling efficiency, we directly learn the embedding to generate images at target resolutions of 128x128 for low-resolution datasets (e.g., CIFAR10/100, MedMNISTv2) and 256x256 for other datasets. This modification significantly reduces image generation time by 27x and 6.5x for 128x128 and 256x256 settings, respectively, making our method more suitable for data augmentation. We summarize the total runtime (i.e., embedding learning and sampling) in Figure~\ref{fig:runtime} and provide a detailed runtime analysis in Appendix~\ref{app:runtime}. We evaluate all the models on real images and we resize all generated images to match the original real images' resolution, ensuring a fair comparison. Additional details are provided in Appendix ~\ref{app:imple} for conciseness. 
% Our code is available in the Supplementary Material.
Our code is available at \url{https://github.com/yongchao97/diffusion_inversion}.

\subsection{Generator Quality and Data Size Matter}
\begin{wrapfigure}{r}{0.46\textwidth}
  \centering
  % \hspace{-0.05in}
  \vspace{-0.2in}
  \includegraphics[width=1.0\linewidth]{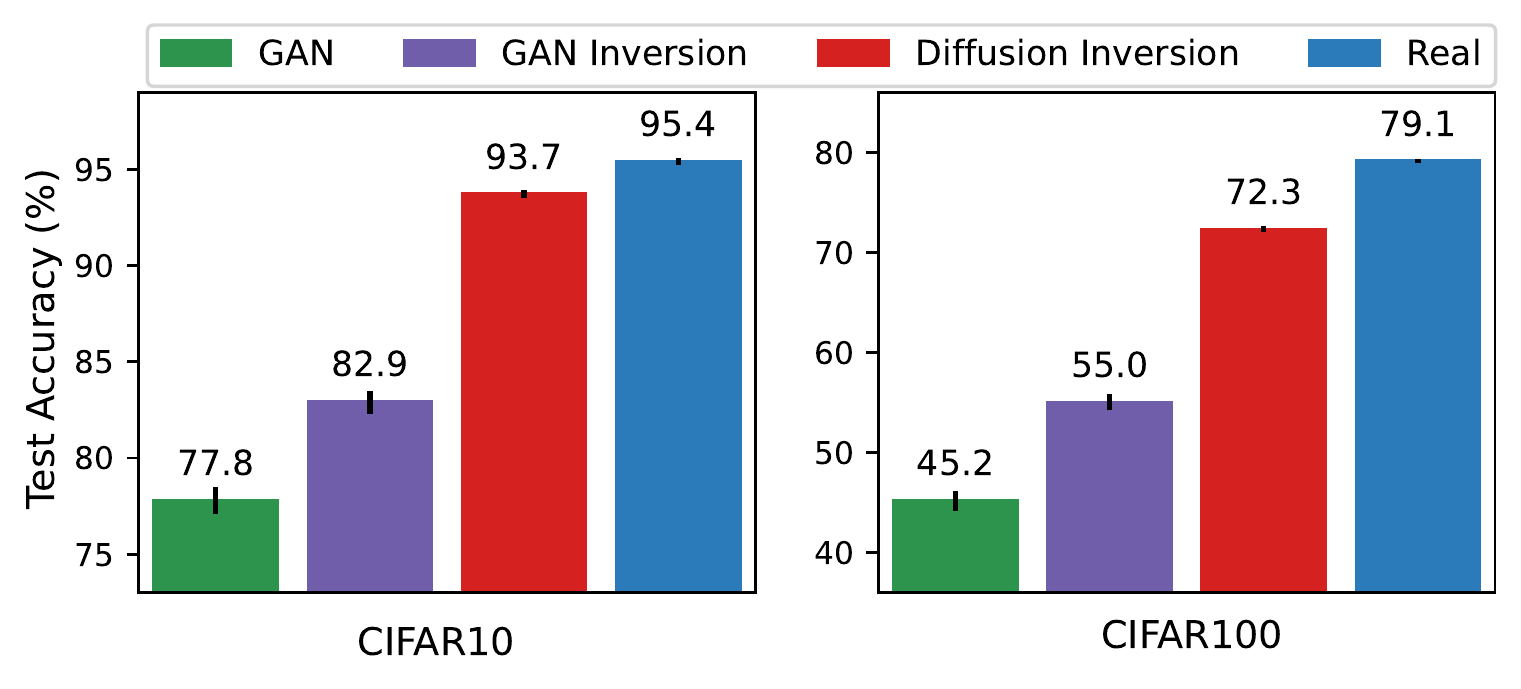}
  \vspace{-0.25in}
  \caption{\footnotesize{Our method outperforms both GAN and GAN Inversion techniques when trained on datasets of equivalent size to the original real dataset, highlighting the significance of a high-quality pre-trained generator.}}
  \label{fig:compare2gan}
  \vspace{-0.15in}
\end{wrapfigure}

\paragraph{Generator Quality} To investigate the influence of generator quality on producing high-quality datasets for downstream classifier training, we initially compare our approach to the GAN Inversion method~\citep{abdal2019image2stylegan} on CIFAR10/100. Using a pre-trained BigGAN model from~\citep{zhao2022synthesizing}, trained with a state-of-the-art strategy~\citep{zhao2020differentiable}, we generate three synthetic datasets, each containing 50K examples, equivalent to the original dataset size. The datasets are created using random latent vectors, GAN Inversion, and our method. To evaluate the datasets' quality, we train a ResNet18 on each and report the mean and standard deviation of the test accuracy using five random seeds.

Figure~\ref{fig:compare2gan} highlights the exceptional performance of our method compared to GAN techniques, illustrating that our approach retains more information from the original dataset and that a high-quality pre-trained generator is crucial for producing top-notch datasets for discriminative models. Nevertheless, a gap remains between our method and the original real dataset for fixed, equal-sized datasets, suggesting that information is denser in the real dataset than in the synthetic one.

\paragraph{Scaling in Number of Real Data} Next, we assess the scalability of our approach by evaluating its benefits for downstream classifier training using four datasets. We generate a sufficient number of synthetic images and learn embeddings from real datasets with varying numbers of training examples. For each embedding, we create 45 unique variants and train a ResNet18 on derived datasets.

\begin{figure}[t]
\vspace{-0.1in}
\centering
\begin{subfigure}[b]{0.245\textwidth}
\centering
\includegraphics[width=1.05\linewidth]{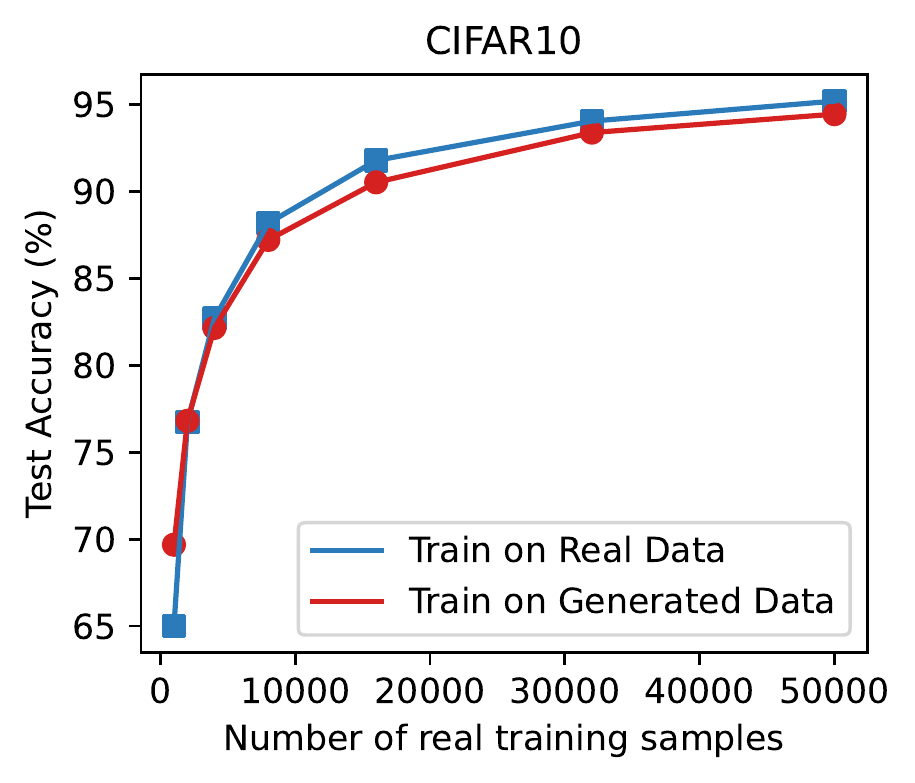}
\end{subfigure}
\hfill
\begin{subfigure}[b]{0.245\textwidth}
\centering
\includegraphics[width=1.05\linewidth]{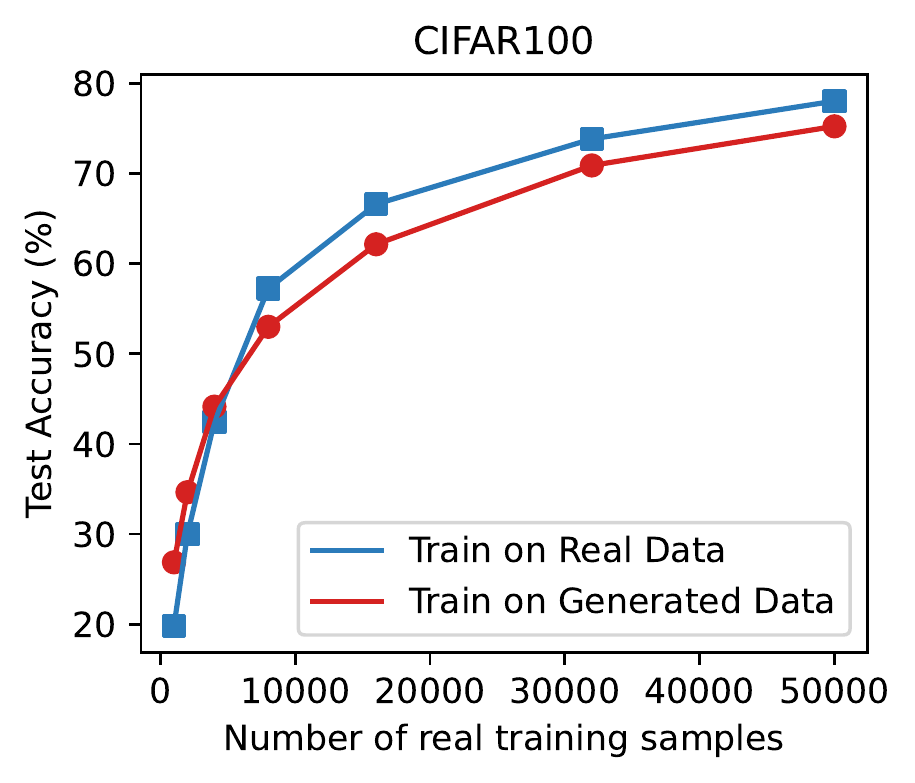}
\end{subfigure}
\hfill
\begin{subfigure}[b]{0.245\textwidth}
\centering
\includegraphics[width=1.05\linewidth]{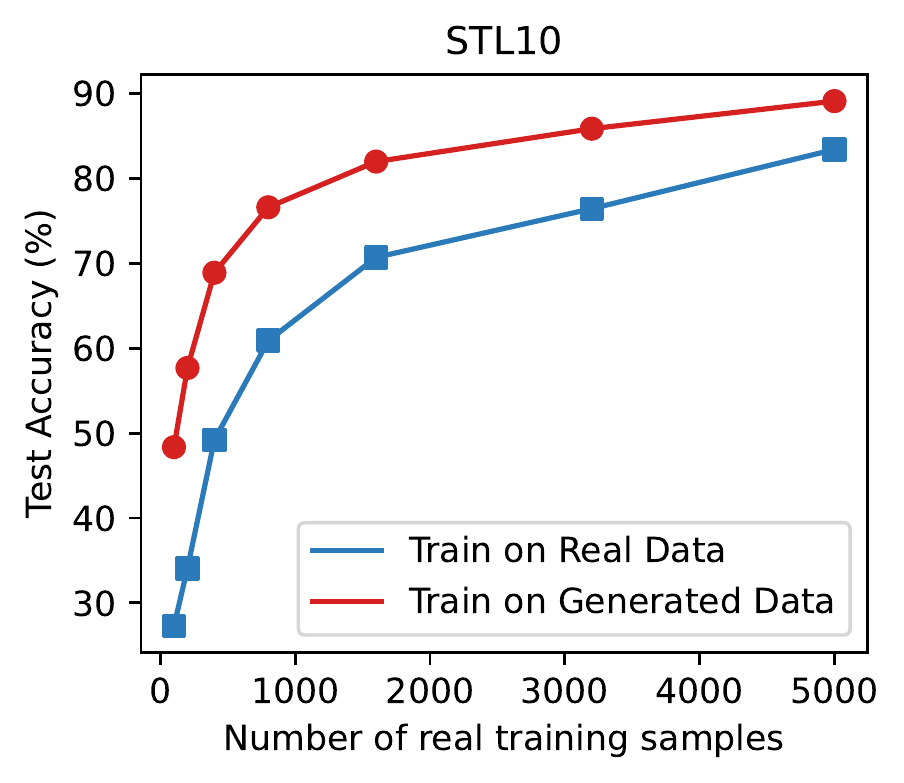}
\end{subfigure}
\hfill
\begin{subfigure}[b]{0.245\textwidth}
\centering
\includegraphics[width=1.05\linewidth]{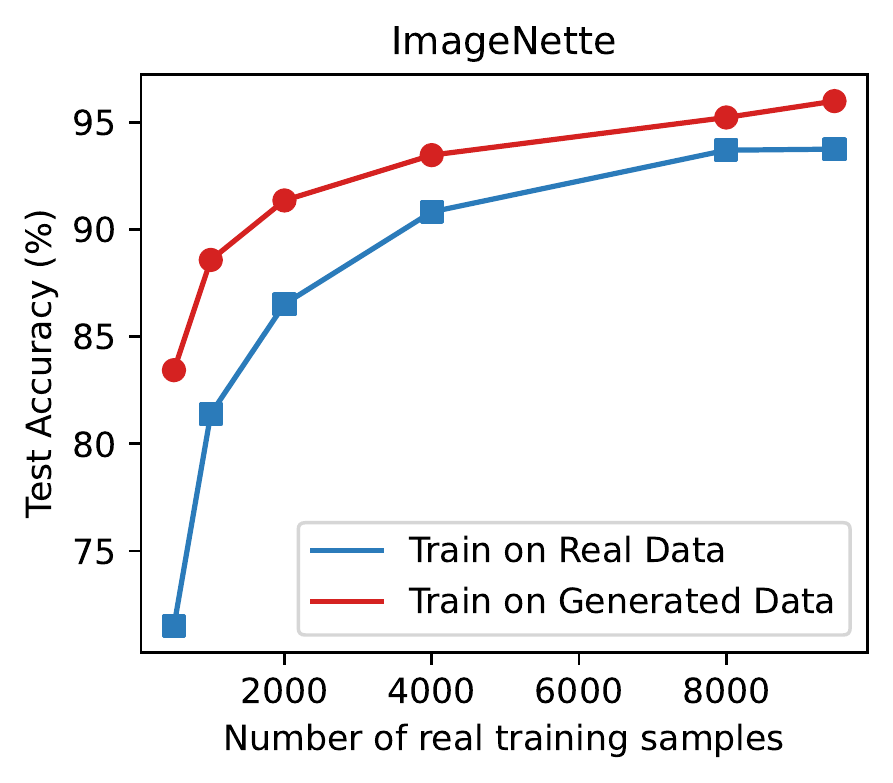}
\end{subfigure}
\caption{\footnotesize{Performance in Relation to Number of Real Data Points. Our approach demonstrates substantially improved performance in low-data scenarios across all datasets. In high-data scenarios, it exhibits comparable performance for low-resolution datasets and superior performance for high-resolution datasets.}}
\vspace{-0.15in}
\label{fig:scaling}
\end{figure}

Figure~\ref{fig:scaling} demonstrates that our method outperforms real data in low data regimes (2K for CIFAR10 and 4K for CIFAR100) for low-resolution datasets like CIFAR10/100 but is slightly worse when more real training data is available. Conversely, for high-resolution datasets such as STL10 and ImageNette, our method consistently surpasses real data by a significant margin. For example, it improves test accuracy on STL10 from 83.3 to 89.0 and on Imagenette from 93.8 to 95.4. Our method also achieves the same accuracy using 2-3x less real data. Furthermore, Table~\ref{tab:vae_sumamry} indicates that reconstructing test data with the Stable Diffusion model's autoencoder often enhances test accuracy for models trained on synthetic data, as also noted in \citet{razavi2019generating}.
% We summarize the performance using the entire real dataset in Table~\ref{tab:vae_sumamry}, where the model attains better test accuracy on the VAE-processed test data for 3 out of 4 datasets.

\begin{wraptable}{r}{0.47\textwidth}
    \footnotesize
    \centering
    \vspace{-0.10in}
    \caption{\footnotesize{Test accuracy of ResNet18 trained on the VAE-Processed data. Autoencoding results in a substantial loss of information, making it difficult to surpass the performance of the real dataset.}}
    \label{tab:vae}
    \begin{tabular}{ccc}
    \toprule
     & CIFAR10 & CIFAR100 \\
    \midrule
    Real (Original) & $95.1 \pm 0.0$ & $77.9 \pm 0.4$ \\
    Real (32 $\xrightarrow{}$ 64) & $91.4 \pm 0.3$ & $65.5 \pm 0.6$  \\
    Real (32 $\xrightarrow{}$ 128) & $92.5 \pm 0.2$ & $66.2 \pm 0.4$ \\
    Real (32 $\xrightarrow{}$ 256) & $93.4 \pm 0.1$ & $69.8 \pm 0.3$ \\
    Real (32 $\xrightarrow{}$ 512) & $93.5 \pm 0.2$ & $71.1 \pm 0.3$ \\
    \midrule
    Diffusion Inversion & $\mathbf{94.6} \pm \mathbf{0.1} $ & $\mathbf{74.4} \pm \mathbf{0.3}$ \\
    \bottomrule
    \end{tabular}
    % \vspace{-0.04in}
\end{wraptable}

The subpar performance in CIFAR can be attributed to information loss during the autoencoding process. To assess this, we generated four CIFAR10 variants with images autoencoded at different resolutions, starting at 64x64—the Stable Diffusion model's minimum requirement. Table~\ref{tab:vae} demonstrates consistent performance enhancement as resolution increases. However, even at the highest resolution of 512, it still underperforms compared to training on original images. This implies significant information loss during autoencoding or a distribution shift between reconstructed and real images. Our method substantially improves test accuracy on CIFAR10 and CIFAR100 compared to the 128-resolution setting, increasing from 92.5 and 66.2 to 94.6 and 74.4, respectively. Additionally, Table~\ref{tab:real_gen} shows that merging real and generated data can slightly enhance the model's performance.

\paragraph{Scaling in Number of Synthetic Data} We also explore the case where we learn embeddings for every data point in the dataset and continue generating more data. As demonstrated in Figure~\ref{fig:workflow}(Right), increased data consistently improves downstream classifier performance, surpassing the real dataset when roughly three times more data is generated. This scaling trend indicates that extended training time and online data generation could further enhance model performance.

\subsection{Data Distribution and Data Coverage Matter}
\begin{wraptable}{r}{0.52\textwidth}
    \footnotesize
    \centering
    \vspace{-0.15in}
    \caption{\footnotesize{Our method outperforms LECF in all metrics, suggesting that while selecting the optimal threshold enhances baseline LECF outcomes, our approach excels in generating higher quality and more diverse images.}}
    \label{tab:lecf}
    \begin{tabular}{ccccc}
    % \begin{tabular}{C{2.1cm}C{1.0cm}C{1.0cm}C{1.0cm}C{1.0cm}C{1.0cm}C{1.0cm}C{1.0cm}C{1.0cm}}
    \toprule
    Name & FID & Precision & Recall & Coverage\\
    \midrule
    % LECF\_0.0 & 40.852 & 0.552 & 0.415 & 0.431\\
    LECF\_0.95 & 33.6 & 0.648 & 0.392 & 0.486\\
    DI (Ours) & 17.7 & 0.831 & 0.661 & 0.787\\
    \bottomrule
    \end{tabular}
    \vspace{-0.15in}
\end{wraptable}
\paragraph{Comparison against Generic Prompt-Based Steering Methods}
The recent study, Language Enhancement with Clip Filtering (LECF) by~\citet{he2022synthetic}, employs Stable Diffusion to generate data for discriminative models, demonstrating cutting-edge performance in few-shot learning. We compared our approach to LECF in two distinct settings: few-shot learning on EuroSAT~\citep{helber2019eurosat} (where their method showed the most significant performance enhancement) and standard training on STL10.

\begin{figure}[t]
\centering
\begin{subfigure}[b]{0.245\textwidth}
\centering
\includegraphics[width=1.0\linewidth]{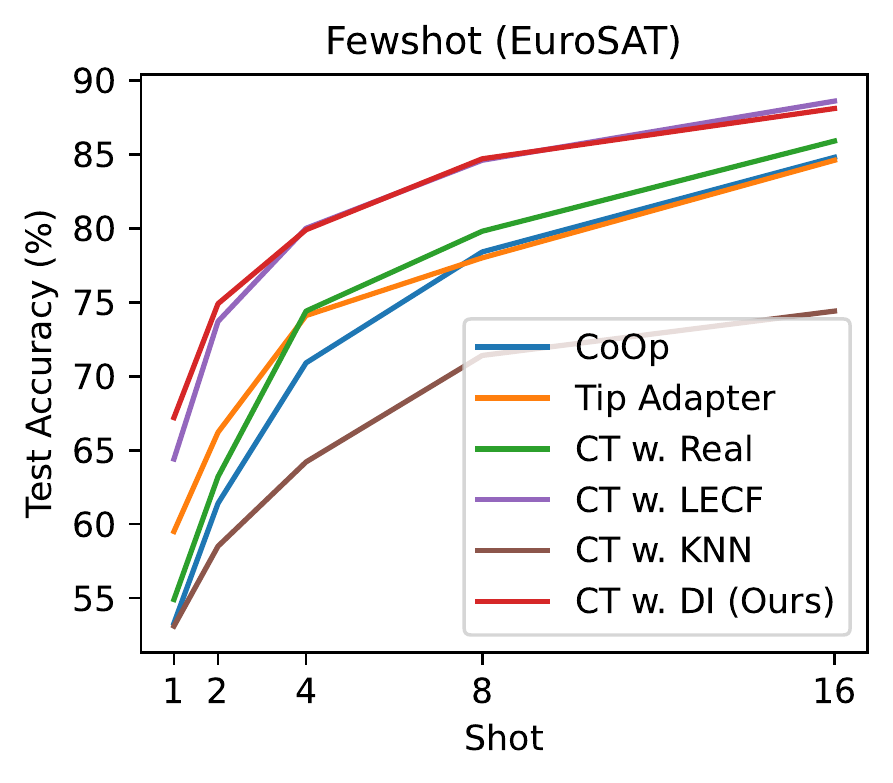}
\caption{Fewshot Learning}
\end{subfigure}
\hfill
\begin{subfigure}[b]{0.245\textwidth}
\centering
\includegraphics[width=1.0\linewidth]{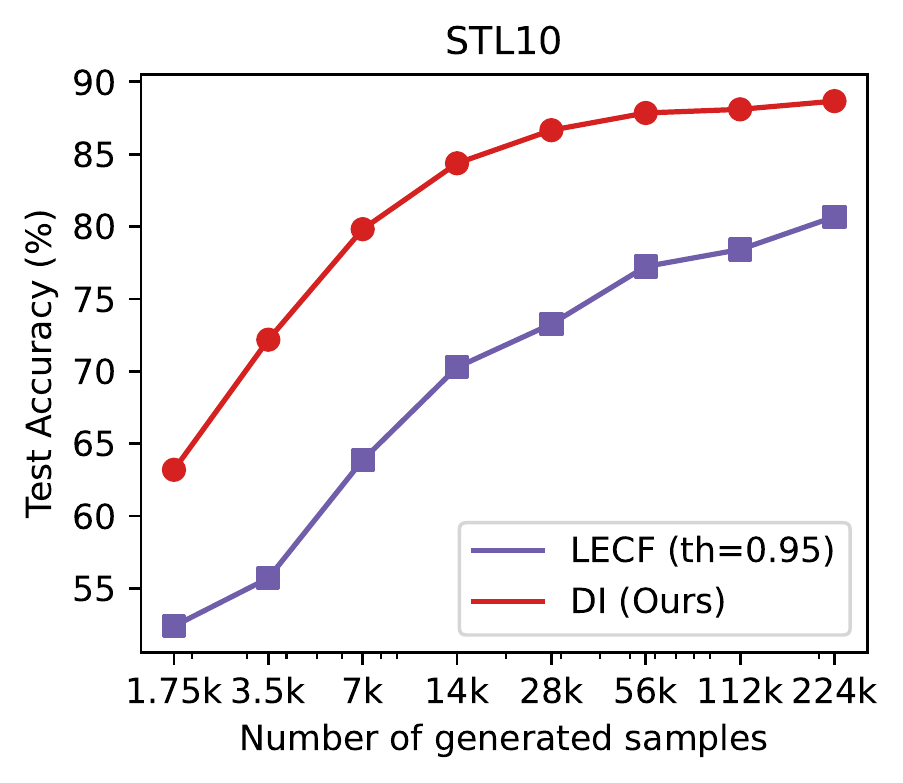}
\caption{Compare to LECF}
\end{subfigure}
\hfill
\begin{subfigure}[b]{0.49\textwidth}
\centering
\includegraphics[width=1.0\linewidth]{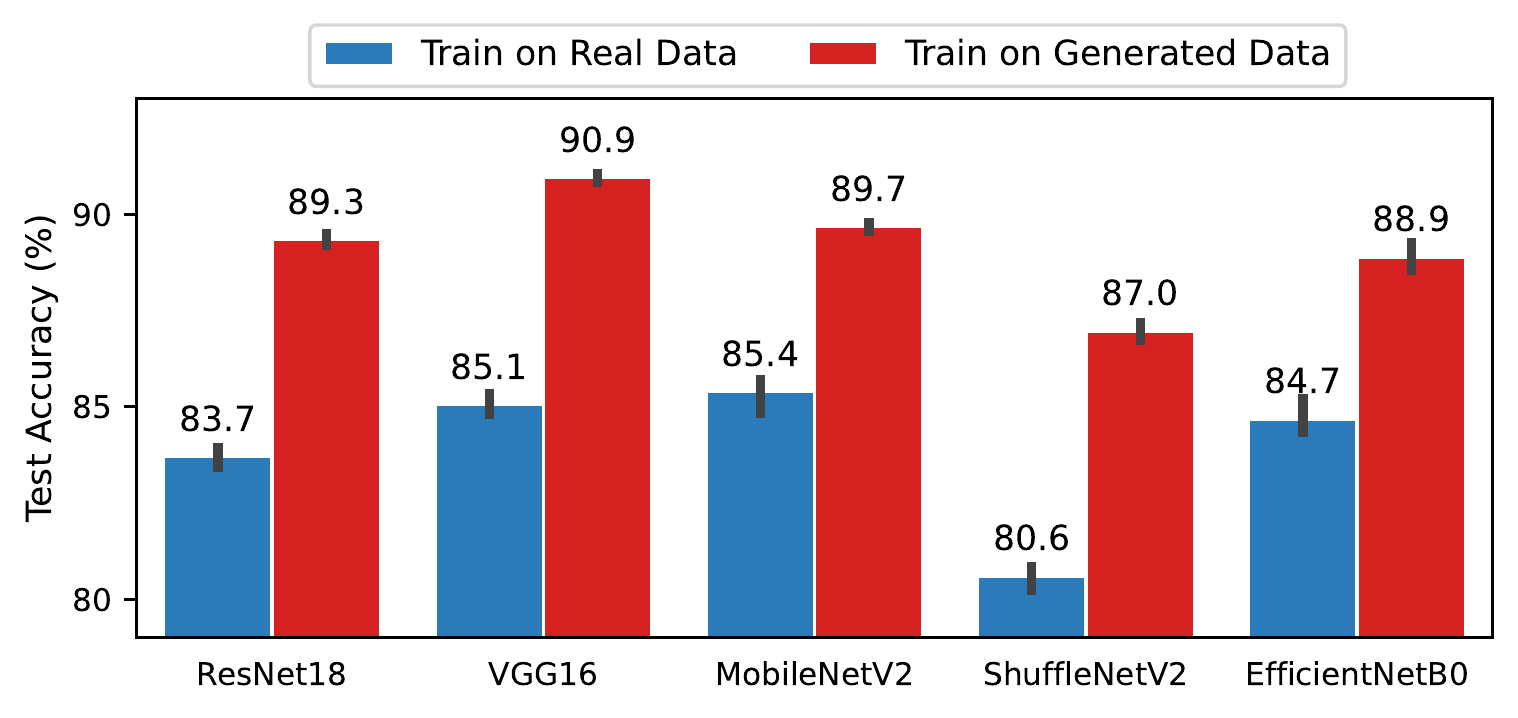}
\caption{Evaluation on Various Architectures}
\end{subfigure}
\vspace{-0.05in}
\caption{\footnotesize{(a) Our method improves few-shot learning performance, yielding results similar to LECF. (b) Our method demonstrates significantly better scalability than LECF on STL10 using only generated data. (c) The synthetic dataset significantly boosts the performance of diverse neural architectures on the STL10 dataset.}}
\label{fig:arch}
\vspace{-0.15in}
\end{figure}

We evaluated our method against CoOP~\citep{zhou2022learning}, Tip Adapter~\citep{zhang2022tip}, and Classifier Tuning (CT) with Real Data~\citep{he2022synthetic} on the EuroSAT dataset. As depicted in Figure~\ref{fig:arch}a, our approach enhances few-shot learning performance, achieving results comparable to LECF. For the STL10 dataset, we analyzed test accuracy progression concerning the number of generated data points. Training a ResNet18 exclusively on generated data and adjusting the Clip Filtering strength of LECF within [0.0, 0.1, 0.3, 0.5, 0.7, 0.9, 0.95, 0.97], we determined that 0.95 yielded optimal performance. Figure~\ref{fig:arch}b highlights our method's superior scaling capabilities compared to LECF, owing to its consideration of domain shifts and improved coverage. This is further evidenced in Table~\ref{tab:lecf} and~\ref{tab:lecf_app}, where we compare FID, precision, recall, and coverage between our method and LECF.

\paragraph{Comparison with KNN Retrieval from LAION}
Stable Diffusion, trained on the LAION dataset \citep{schuhmann2022laion}, prompted the question of synthetic dataset necessity versus similar image retrieval for data augmentation. We assessed this using KNN retrieval from LAION with clip retrieval on the STL10 dataset. Test accuracies achieved were 85.4\%, 88.4\%, and 90.9\% for k=10, 25, and 50, respectively. Our method generated 88.7\% test accuracy with 45 data points per embedding, slightly surpassing 25-image retrieval but not 50-image retrieval. This suggests KNN retrieval as a strong baseline for well-represented target classes like airplanes, cars, and dogs in Stable Diffusion training distribution.

\begin{wraptable}{r}{0.52\textwidth}
    \footnotesize
    \centering
    \vspace{-0.05in}
    \caption{\footnotesize{Comparing our method with the KNN retrieval baseline on LAION-5B, our approach consistently excels across three medical imaging datasets, emphasizing its proficiency in handling distribution shifts and data coverage.}}
    \label{tab:biology}
    \begin{tabular}{ccccc}
    \toprule
     & K=10 & K=25 & K=50 & DI (Ours)\\
    \midrule
    % STL10  & 85.4 & 88.4 & $\textbf{90.9}$ & 88.7\\
    PathMNIST & 22.5 & 29.9 & 23.4 & $\textbf{82.1}$\\
    DermaMNIST & 23.0 & 27.8 & 22.1 & $\textbf{67.5}$\\
    BloodMNIST & 21.7 & 27.7 & 25.8 & $\textbf{93.7}$\\
    \bottomrule
    \end{tabular}
    % \vspace{-0.2in}
\end{wraptable}

However, we argue that this method falls short when significant distribution shifts occur between target and source domains, especially in specialized fields like medical imaging. To demonstrate this, we analyzed three distinct MedMNISTv2 datasets: PathMNIST, DermaMNIST, and BloodMNIST. As depicted in Table~\ref{tab:biology}, our approach consistently surpasses the KNN retrieval baseline. It is crucial to acknowledge that LECF would falter in this scenario due to significant distribution shifts and challenges in creating effective prompts. Additionally, KNN retrieval does not improve few-shot learning performance on EuroSAT, as demonstrated in Figure~\ref{fig:arch}a. The high-quality generated images for these specialized domain datasets, depicted in Figure~\ref{fig:syn_data_vis}, closely resemble the original dataset, underscoring the importance of a steering method that addresses distribution shift and data coverage. However, Table~\ref{tab:real_gen} reveals that the generated data underperforms compared to real data, primarily due to the dataset's low resolution. Consequently, a substantial amount of information is 
lost during the autoencoding process, as observed in the CIFAR case.

\subsection{Evaluation on Various Architectures}
The value of synthetic data increases significantly when compatible with various neural architectures. To assess the effectiveness of our generated data, we tested its performance on a diverse set of popular neural network architectures, including ResNet18~\cite{he2016deep}, VGG16~\cite{simonyan2014very}, MobileNetV2~\cite{sandler2018mobilenetv2}, ShuffleNetV2~\cite{ma2018shufflenet}, and EfficientNetB0~\cite{tan2019efficientnet}, using the STL10 dataset. As depicted in Figure \ref{fig:arch}c, synthetic images notably enhance performance across all examined architectures, indicating that our method successfully extracts generalizable knowledge from the pre-trained Stable Diffusion and incorporates it into the generated dataset.

\subsection{Comparison against Image Data Augmentation Methods}
We assess our approach by comparing it to widely-adopted data augmentation techniques for STL10 image classification. These encompass standard methods such as AutoAugment~\citep{cubuk2018autoaugment}, RandAugment~\citep{cubuk2020randaugment}, and CutOut~\citep{devries2017improved}; interpolation-based techniques like MixUp~\citep{zhang2017mixup}, CutMix~\citep{yun2019cutmix}, and AugMix~\citep{hendrycks2019augmix}; as well as the adversarial domain augmentation (ADA) method ME-ADA~\citep{zhao2020maximum}. A detailed overview of each technique can be found in Appendix \ref{imp:data_aug}.

Table \ref{tab:aug} demonstrates that our method (89.5\%) combined with default data augmentation (i.e., random crop and flip) surpasses all previously mentioned techniques (as seen in the first row). Additionally, our approach effectively complements other data augmentation methods, and their combined implementation can yield even better performance.

\begin{table*}[t]
    \centering
    \caption{\footnotesize{Data Augmentation Techniques on STL10: Our approach, combined with default augmentation (crop and flip), consistently outperforms alternatives and can be further improved by merging with other techniques.}}
    \label{tab:aug}
    % \begin{tabular}{p{2.1cm}p{0.9cm}p{1cm}p{1cm}p{1cm}p{1cm}p{1cm}p{1cm}p{1.3cm}}
    \footnotesize
    \begin{tabular}{C{2.1cm}C{1.0cm}C{1.0cm}C{1.0cm}C{1.0cm}C{1.0cm}C{1.0cm}C{1.0cm}C{1.0cm}}
    \toprule
     & Default & Auto-Aug & Rand-Aug & CutOut & MixUp & CutMix & AugMix & ME-ADA\\
    \midrule
    Original Dataset & 83.2 & 87.0 & 86.3 & 84.3 & 89.4 & 88.1 & 83.8 & 83.4\\
    Synthetic (Ours) & $\textbf{89.5}$ & $\textbf{91.5}$ & $\textbf{91.0}$ & $\textbf{89.5}$ & $\textbf{91.5}$ & $\textbf{92.6}$ & $\textbf{89.2}$ & $\textbf{89.1}$\\    
    \bottomrule
    \end{tabular}
    \vspace{-0.05in}
\end{table*}

\begin{figure}[t]
% \vspace{0.1in}
\begin{subfigure}[b]{0.245\textwidth}
\centering
\includegraphics[width=1.0\linewidth]{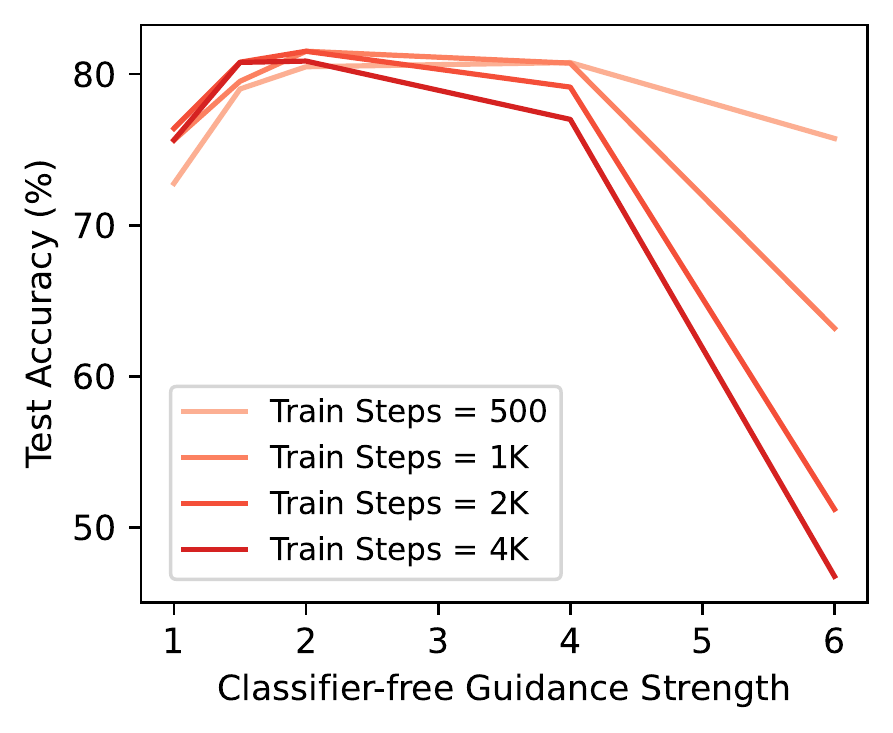}
\caption{Classifier-free Guidance}
\end{subfigure}
\hfill
\begin{subfigure}[b]{0.245\textwidth}
\centering
\includegraphics[width=1.0\linewidth]{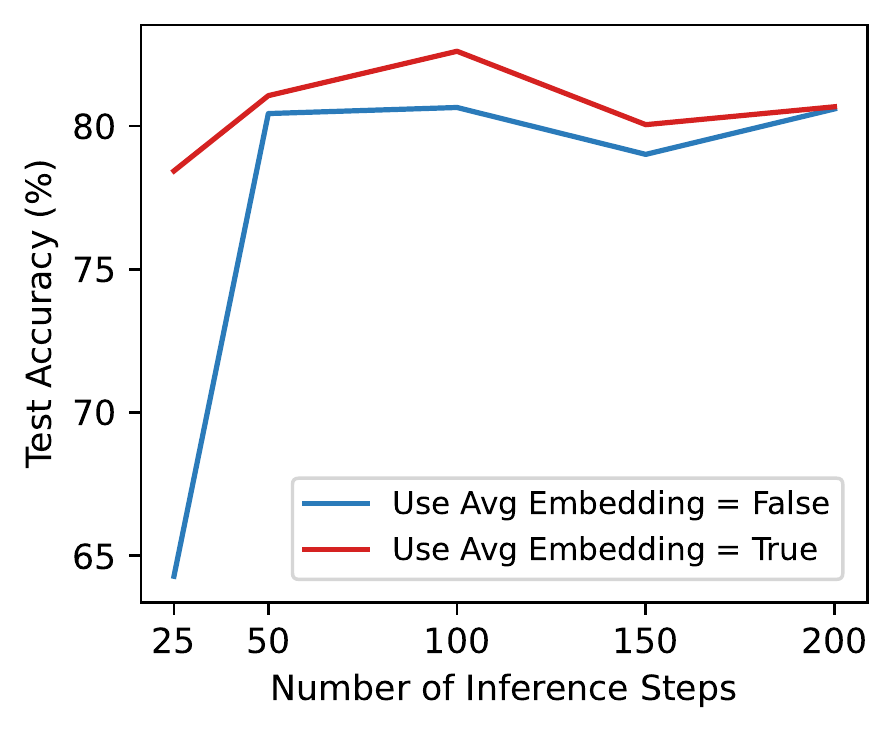}
\caption{Inference Steps}
\end{subfigure}
\hfill
\begin{subfigure}[b]{0.245\textwidth}
\centering
\includegraphics[width=1.0\linewidth]{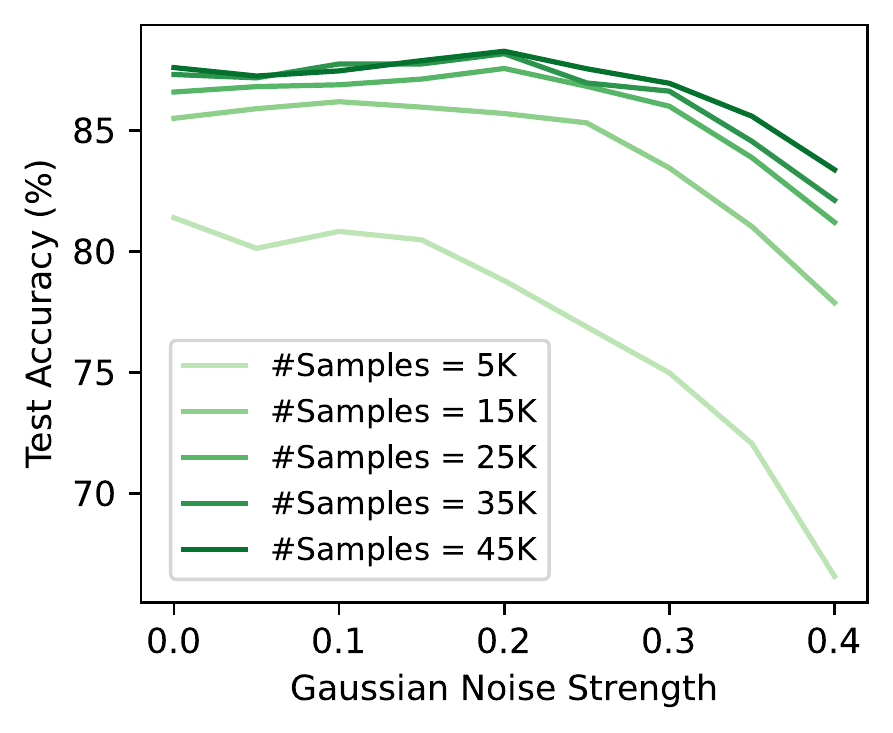}
\caption{Gaussian Noise}
\end{subfigure}
\hfill
\begin{subfigure}[b]{0.245\textwidth}
\centering
\includegraphics[width=1.0\linewidth]{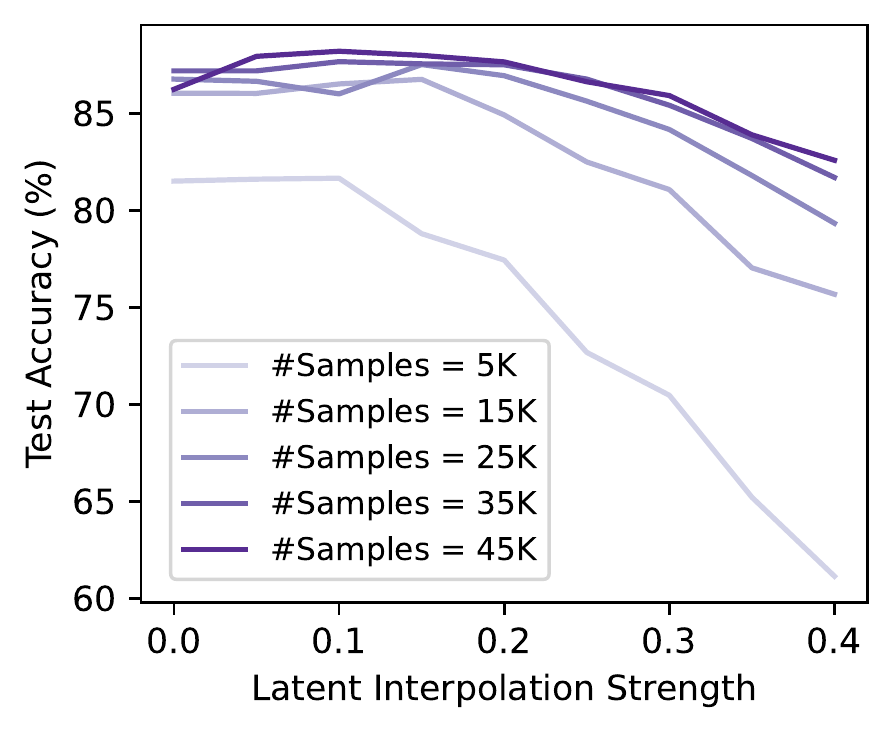}
\caption{Latent Interpolation}
\end{subfigure}
\caption{\footnotesize{(a) Impact of training steps and classifier-free guidance strength on model performance. (b) Influence of inference steps and unconditional embedding on model performance. (c) Performance variation with Gaussian noise as data generation increases. (d) Latent interpolation effects on performance during data generation.}}
\label{fig:tsteps}
\vspace{-0.1in}
\end{figure}

\section{Quantitative Analysis}
We perform extensive quantitative evaluations on STL10 to assess the impact of various design choices and hyperparameters, including the number of training and inference steps, classifier-free guidance strength, unconditional embedding, latent interpolation, and Gaussian noise. In each experiment, we only modify the target hyperparameter while keeping all others at their default values. 

\subsection{Training Steps and Classifier-free Guidance Strength}
Figure~\ref{fig:tsteps}a illustrates the performance correlation with increasing training steps for embedding vectors and classifier-free guidance strength. Results reveal marginal performance enhancement beyond 1K steps, and optimal guidance strength diminishes with extensive training. A high guidance strength leads to a significant performance drop. In practice, training for 2K steps and employing a guidance strength between 2 and 4 serves as a suitable starting point.

\subsection{Inference Steps and Unconditional Embedding}\label{sec:avgemb}
Figure~\ref{fig:tsteps}b demonstrates that utilizing the mean of all learned vectors as class-conditioning input for unconditional models consistently surpasses the performance of the text encoder's output with an empty string. We also note that the learned image embedding does not effectively generate images of different resolutions. For instance, a latent vector learned from a 512-resolution image struggles to produce 128-resolution images. This emphasizes the inadequacy of the empty string embedding, as the initial text encoder is co-trained with the denoising model on higher-resolution images (512x512). Regarding inference steps, we determine that 100 steps provide an optimal balance between performance and computational cost, making it our default choice.

\begin{figure*}[t]
\begin{center}
\vspace{-0.15in}
\centerline{\includegraphics[width=1.0\linewidth]{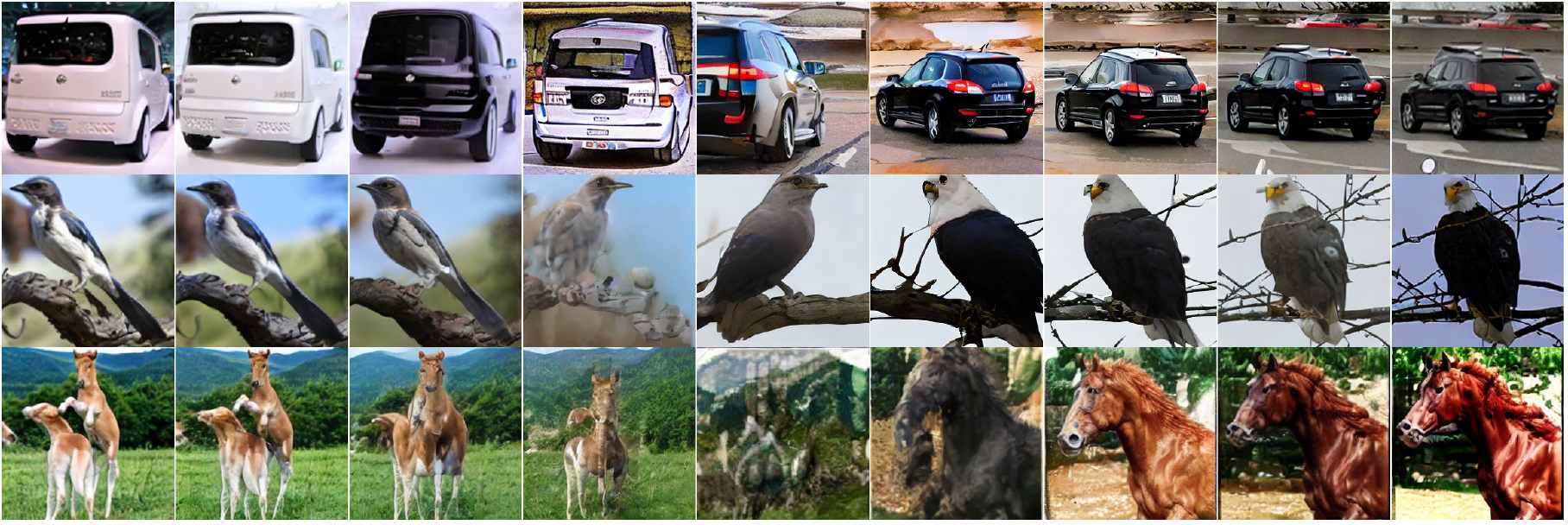}}
\vspace{-0.1in}
\caption{\footnotesize{Image variants generated by interpolating two embedding vectors with varying interpolation strengths ($\alpha$): 0.0, 0.1, 0.2, 0.3, 0.5, 0.7, 0.8, 0.9, 1.0 (left to right). Some vector pairs produce novel, natural images at any strength, while others require low strengths.}}
\label{fig:ab_latent_vis}
\end{center}
\vspace{-0.3in}
\end{figure*}

\subsection{Latent Interpolation and Gaussian Noise}\label{sec:gauss_latent}

\paragraph{Gaussian Noise} We examine the impact of Gaussian noise on model performance by altering its strength and setting the latent interpolation strength $\alpha$ to $0$. Figure~\ref{fig:tsteps}c shows the correlation between Gaussian noise strength and model test accuracy. Our results suggest that optimal performance is obtained when creating a dataset of equal size to the original without noise perturbation. However, when sampling additional data, it is beneficial to raise the noise strength accordingly, starting with a noise strength of $\lambda=0.2$. We also provide a summary of various metrics, such as FID, precision, recall, and coverage in Table~\ref{tab:noise}, and depict the generated images at different noise levels in Figure~\ref{fig:ab_gaussian_vis}.

\begin{wraptable}{r}{0.5\textwidth}
    \footnotesize
    \centering
    \vspace{-0.1in}
    \caption{\footnotesize{Increasing interpolation strength adversely affects FID and precision while improving recall. Coverage peaks around an interpolation of 0.1, indicating a trade-off between generation quality and diversity.}}
    \label{tab:latent_interp}
    \begin{tabular}{ccccc}
    \toprule
     alpha & FID & Precision & Recall& Coverage\\
    \midrule
    0.00 & 17.9 & 0.894 & 0.644 & 0.753\\
    0.10 & 17.7 & 0.831 & 0.661 & 0.787\\
    0.20 & 26.2 & 0.635 & 0.751 & 0.739\\
    0.30 & 43.2 & 0.448 & 0.787 & 0.605\\
    0.40 & 62.8 & 0.328 & 0.805 & 0.500\\
    \bottomrule
    \end{tabular}
    \vspace{-0.1in}
\end{wraptable}

\paragraph{Latent Interpolation} In this study, we examine the impact of latent interpolation on model performance by varying interpolation strength and setting Gaussian noise strength ($\lambda$) to 0. Figure~\ref{fig:tsteps}d illustrates the relationship between interpolation strength and performance, highlighting that high strength significantly diminishes performance. Unlike Gaussian noise strength, a larger sample size does not benefit high strength, with the optimal value lying between 0.1 and 0.15. Figure \ref{fig:ab_latent_vis} displays samples, indicating that unique and realistic images can be generated at any interpolation strength, provided the two embedding vectors are highly compatible. However, poorly chosen embeddings can produce perplexing images at $\alpha=0.3$ (i.e., third row). Table \ref{tab:alpha} demonstrates how FID, precision, recall, density, and coverage vary with interpolation strength $\alpha$. In our experiment, we randomly select two embedding vectors to generate new images, resulting in a small optimal interpolation strength. We propose that a more refined algorithm for selecting embeddings could further enhance performance. 

\section{Conclusion}
We present Diffusion Inversion, a simple yet effective method for generating high-quality synthetic data that enhances image classification by utilizing pre-trained generative models. Our approach surpasses generic prompt-based steering methods and KNN retrieval from LAION-5B, seamlessly integrating with standard augmentation techniques and benefiting various neural architectures, substantially improving few-shot learning performance. This research emphasizes the potential of employing pre-trained generative models for data augmentation, particularly in specialized domains with significant data distribution shifts or costly data acquisition and curation.

\textbf{Limitations \& Societal Impacts} Although our method significantly reduces total generation time (Figure~\ref{fig:compare2gan}a), scaling it to large-scale datasets like ImageNet~\citep{russakovsky2015imagenet} presents challenges due to substantial storage requirements and inefficient sampling of Stable Diffusion. Incorporating fast sampling techniques~\citep{meng2022distillation} represents a promising direction for maximizing the impact of diffusion-based data generation for discriminative models. Moreover, adopting this approach in real-life applications may necessitate careful consideration; we discuss the societal impact of using these models in Appendix~\ref{sec:societal}.

\begin{ack}
We would like to thank Keiran Paster, Silviu Pitis, Harris Chan, Yangjun Ruan, Michael Zhang, Leo Lee, and Honghua Dong for their valuable feedback. Jimmy Ba was supported by NSERC Grant [2020-06904], CIFAR AI Chairs program, Google Research Scholar Program and Amazon Research Award. Resources used in preparing this research were provided, in part, by the Province of Ontario, the Government of Canada through CIFAR, and companies sponsoring the Vector Institute for Artificial Intelligence.
\end{ack}

\bibliographystyle{unsrtnat}
\bibliography{neurips_2023}

\clearpage
\newpage
\appendix

\newpage
\section{Societal Impact }\label{sec:societal}
As generative models advance, harnessing them for high-quality training data can substantially cut time and resources spent on data collection and annotation. Our method offers a streamlined, efficient means of utilizing these powerful models, potentially allowing smaller organizations and researchers with limited resources to develop effective machine learning models more feasibly.

However, implementing this approach in real-life applications requires caution due to concerns about bias and fairness~\citep{scheuerman2021datasets}. Generative models, such as Stable Diffusion~\citep{rombach2022high}, are trained on extensive, diverse, and uncurated internet data that may contain harmful biases and stereotypes~\citep{bender2021dangers, birhane2021multimodal}. These biases can worsen during generation~\citep{cho2022dall}, leading to discriminatory AI decision-making. However, our method can be utilized to generate diverse, high-quality data for underrepresented groups, fostering fairer and less biased AI systems.

Another potential drawback is the misuse of generated data. High-quality generated data could be exploited for malicious purposes, such as deepfakes~\citep{lyu2020deepfake}, leading to the proliferation of misinformation and manipulation in various domains, including politics, social media, and entertainment.

To counter these negative societal impacts, it is vital to ensure responsible development and deployment of the Diffusion Inversion method and related technologies. This entails incorporating mechanisms to detect and mitigate biases, exploring ethical policies and regulations for synthetic data use, and conducting further research to curate generated data and create fairer multimodal representations of the real world. Establishing responsible practices and guidelines for such methods is crucial for promoting their positive societal impact.

\section{Experimental Details}\label{app:imple}
\subsection{Implementation Details}\label{app:imple_detail}
\paragraph{Datasets} We evaluate our methods on the following datasets:
i) \textbf{CIFAR}~\citep{krizhevsky2009learning}: A standard image dataset with two tasks, CIFAR10 (10 classes) and CIFAR100 (100 classes), each containing 50,000 training examples and 10,000 test examples at a 32x32 resolution. ii) \textbf{STL10}~\citep{coates2011analysis}: An image dataset of 113,000 color images at a 96x96 resolution, designed for semi-supervised learning. It has 5,000 labeled training images and 8,000 labeled test images across ten classes. We use only the labeled portion to test our method's performance on higher-resolution, low-data settings. iii) \textbf{ImageNette}~\citep{imagenette}: A 10-class subset of ILSVRC2012~\citep{russakovsky2015imagenet} containing 9,469 training and 3,925 testing examples, resized to a 256x256 resolution. iv) \textbf{EuroSAT}~\citep{helber2019eurosat}: A dataset based on Sentinel-2 satellite images, covering 13 spectral bands and consisting of 27,000 labeled and geo-referenced samples across ten classes. v) \textbf{MedMNISTv2}~\citep{yang2023medmnistv2}: A large-scale collection of standardized biomedical images, including 12 datasets pre-processed into 28x28 resolution. We use three datasets focused on multi-class image classification: PathMNIST, DermaMNIST, and BloodMNIST. We only learn the embedding for 500 images per class from the train split to reduce the computation cost.

\paragraph{Training} We utilize the publicly accessible 1.4 billion-parameter text-to-image model by \citet{rombach2022high}, pretrained on the LAION-2B(en) dataset\footnote{We use the checkpoint ``CompVis/stable-diffusion-v1-4'' from Hugging Face. \url{https://huggingface.co/CompVis/stable-diffusion-v1-4}}. The model's default image resolution is 512x512, with a minimum functional requirement of 64x64. However, some datasets have a 32x32 resolution. To accommodate this and our training budget, we resize CIFAR10/100 and MedMNISTv2 images to 128x128 and STL-10, EuroSAT and ImageNette images to 256x256. We optimize Eq.~\ref{method:loss} using AdamW \citep{loshchilov2017decoupled} with a constant learning rate of 0.03 for up to 3K steps to learn the conditioning vector for each real dataset image, without data augmentation.

\paragraph{Sampling} Although on-the-fly data generation is ideal, it is computationally costly. We pre-generate a fixed-size dataset and train models on it. Unless specified, we generate each new image in 100 denoising steps using 3K-step checkpoints with classifier-free guidance strength of 2, Gaussian noise strength of 0.1, and embedding interpolation strength of 0.1.

\paragraph{Evaluation} In order to assess the quality of the generated synthetic dataset, we employ a ResNet18 model \cite{he2016deep} and train it on both real and synthesized data at the default resolution. Our training approach utilizes a conventional methodology where the ResNet is trained using Stochastic Gradient Descent (SGD) with momentum, featuring a batch size of 128, a cosine learning rate schedule with an initial learning rate of 0.1, and a standard data augmentation scheme comprising random horizontal flips and random crops following zero-padding. However, for MedMNISTv2, we opt for the AdamW optimizer \citep{kingma2014adam} with a learning rate of 1e-3 and a weight decay of 5e-4.

\subsection{Experimental Setups}
\subsubsection{Generator Quality}\label{imp:gan}
To emphasize the significance of generator quality in producing high-quality datasets for discriminative model training, we first compare our approach with the GAN Inversion method (using a pre-trained BigGAN by \citet{abdal2019image2stylegan}) on CIFAR10 and CIFAR100. We learn a latent vector $z \in \mathbf{R}^{d_z}$ for each image $x \in \mathbf{R}^{d_i}$ in the real dataset by minimizing the weighted sum of feature and pixel distances between synthetic and real images, with a pre-trained feature extractor $\psi_{\vartheta}$, feature dimension $d_f$, and default $\lambda_{\text {pixel }}=1$.
\begin{equation*}
\underset{\boldsymbol{z}}{\arg \min } \frac{1}{d_f}\left\|\psi_{\vartheta}(G(\boldsymbol{z}))-\psi_{\vartheta}(\boldsymbol{x})\right\|^2+\frac{\lambda_{\text {pixel }}}{d_I}\|G(\boldsymbol{z})-\boldsymbol{x}\|^2
\end{equation*}

Using the pre-trained BigGAN provided by \citet{zhao2022synthesizing} and trained with a state-of-the-art strategy \cite{zhao2020differentiable}, we create three synthetic datasets equivalent in size to the original dataset. These datasets are generated using random latent vectors, GAN Inversion, and our method with classifier-free guidance of 2 and checkpoints at 3K steps. To evaluate dataset quality, we train a ResNet18 on each dataset and report the mean and standard deviation of five random seeds.

\subsubsection{Scaling in Relation to Real Data Size}\label{imp:scaling_real}
In Figure~\ref{fig:scaling}, we obtain an embedding for each data point and generate 45 samples per embedding over 100 denoising steps. We use checkpoints at 1K, 2K, and 3K steps, a classifier-free guidance strength sampled from [2, 3, 4], and Gaussian noise and embedding interpolation strengths of 0.1.

\subsubsection{Comparison against Image Data Augmentation Methods}\label{imp:data_aug}
In the following section, we outline the data augmentations employed in our experiments. Utilizing the STL10 dataset for all tests, the results can be found in Table \ref{tab:aug}. Our training methodology adheres to the approach delineated in Section \ref{app:imple_detail}. Additionally, we reuse the generated data from Section \ref{imp:scaling_real} for analysis.\\

i) \textbf{AutoAugment~\citep{cubuk2018autoaugment}:} We utilize torchvision.transforms.AutoAugment, PyTorch's built-in implementation of AutoAugment, with the ImageNet policy comprising 25 transforms. During training, one transform is randomly chosen and applied with a specified probability and magnitude.\\
ii) \textbf{RandAugment~\citep{cubuk2020randaugment}:} Similar to AutoAugment, we randomly select two operations from a list of 14 and apply them with certainty.\\
iii) \textbf{CutOut~\citep{devries2017improved}:} Our CutOut implementation masks out a random square region, sized at 1/8 of the input image.\\
iv) \textbf{MixUp~\citep{zhang2017mixup}:} We use interpolated images as new inputs for network training by combining a permuted batch of inputs with the original batch, sampling interpolation strength from the beta distribution (beta = 1). The loss function is adapted accordingly.\\
v) \textbf{CutMix~\citep{yun2019cutmix}:} We replace a region of each input with a corresponding region from another input by permuting each batch and sampling a region size from the beta distribution. The modified loss function from MixUp is used, with lam representing the area ratio of the selected region to the image.\\
vi) \textbf{AugMix~\citep{hendrycks2019augmix}:} Images are augmented and mixed with the original image by sampling and composing operations. One chain is randomly applied to obtain the augmented image, which is then combined with the original image using an interpolation weight sampled from the beta distribution (alpha=1). Our implementation uses PyTorch's torchvision.transforms.AugMix method with default parameters.\\
vii) \textbf{ME-ADA~\citep{zhao2020maximum}:} In ME-ADA, an adversarial data augmentation method, a minimax procedure runs K times. Each cycle consists of a minimization stage (T\_{min} steps of network training) and a maximization stage (converting input-label pairs to adversarial examples by nudging inputs towards the loss function gradient).
\clearpage
\newpage
\section{Additional Results}
\subsection{Run Time Analysis}\label{app:runtime}
Our method consists of two main components: learning embeddings and sampling. For ImageNette and STL-10, we learn an image embedding and prompt the Stable Diffusion model to generate a 256x256 image. Training the embedding for 3,000 steps on an Nvidia A40 GPU takes an average of 84.1 seconds. For CIFAR10/100, we learn an embedding that enables direct generation of 128x128 images, with an average learning time of 18.8 seconds per embedding.

Sampling with the learned embeddings also incurs computational costs. Standard Stable Diffusion sampling takes about 5.28 seconds to generate a 512x512 image with 100 diffusion steps. However, generating a 256x256 or 128x128 image based on the learned embedding takes only 0.82 seconds (6.44 times faster) or 0.20 seconds (26.4 times faster), respectively. This speedup results from the absence of a CLIPText encoder for text prompt embedding and the lower-dimensional diffusion process. Dimension size, rather than the text encoder, plays a more significant role in time reduction. Below are average times for generating one image using 100 inference steps:

- (64, 64, 4) with Text Encoder: 5.28s\\
- (64, 64, 4) without Text Encoder: 5.19s\\
- (32, 32, 4) without Text Encoder: 0.82s\\
- (16, 16, 4) without Text Encoder: 0.20s

To generate 45 samples per learned embedding (our default setting), the total time for both embedding learning and sampling is approximately 121 seconds for ImageNette/STL10 and 27.8 seconds for CIFAR10/100. In comparison, standard Stable Diffusion sampling for 45 images takes 237.6 seconds. Our method is nearly twice as fast for ImageNette/STL10 and 8.5 times faster for CIFAR10/100. Moreover, the amortized cost of learning the embedding decreases when generating more data, making our approach more suitable as a data augmentation tool. We summarize the total runtime (i.e., embedding learning and sampling) in Figure~\ref{fig:runtime}.

\clearpage
\newpage
\subsection{Model Performance}
\paragraph{Model achieves better accuracy on VAE processed test data} As shown in Table~\ref{tab:vae_sumamry}, reconstructing test data with the Stable Diffusion model's autoencoder often enhances test accuracy for models trained on synthetic data, as also noted in \citet{razavi2019generating}.
\begin{table}[ht]
    \centering
    \small
    % \vspace{-0.15in}
    \caption{Test accuracy using the entire dataset. Transforming the test data using VAE can often improve the model performance.}
    \label{tab:vae_sumamry}
    \begin{tabular}{cccc}
    \toprule
    &  & \multicolumn{2}{c}{Synthetic Data}  \\
    \cmidrule(l){3-4}
     & Real Data & Original & VAE-Processed\\
    \midrule
    CIFAR10 & $\textbf{95.1} \pm \textbf{0.0}$ & $94.6 \pm 0.1$ & $94.7 \pm 0.1$ \\
    CIFAR100 & $\textbf{77.9} \pm \textbf{0.4}$ & $74.4 \pm 0.3$ & $75.2 \pm 0.2$   \\
    STL-10 & $83.3 \pm 0.7$ & $\textbf{89.0} \pm \textbf{0.2}$ & $88.8 \pm 0.2$  \\
    ImageNette & $93.8 \pm 0.2$ & $95.4 \pm 0.1$ & $\textbf{95.6} \pm \textbf{0.1}$  \\
    \bottomrule
    \end{tabular}
\end{table}

\paragraph{Combine real and generated data} In our study, we analyze a model trained on a combination of generated data and synthetic data. We employ a straightforward approach to merge the real and synthetic data. Specifically, at each gradient step, we construct a batch using a mixture of synthetic and real data and train the model following the same protocol as when training on either generated data only or real data only (e.g., optimizer, training steps, and batch size). We experiment with varying the real-to-synthetic mixture ratio from [1:7, 1:3, 1:1, 3:1, 7:1] and report the best performance in Table~\ref{tab:real_gen}. Our observations indicate that although generated data underperforms real data on low-resolution datasets such as CIFAR10, CIFAR100, and PathMNIST, combining both types enhances performance, as also observed by \citet{azizi2023synthetic}. However, in some instances, like Imagenette, DermaMNIST, and BloodMNIST, the combination leads to a slight performance decrease compared to using real or generated data alone. A similar observation was made by~\citet{ravuri2019classification} (Fig. 5), where they found that mixing generated samples with real data degrades Top-5 classifier accuracy for almost all models tested. Concurrently,~\citet{azizi2023synthetic} (Table 4) notes that model performance with higher resolution images does not continue to improve with larger amounts of generative data augmentation after a certain point. This may be attributable to the bias in the generated data inherited from the generative models, suggesting that a more sophisticated method for merging real and generated data is necessary. We leave the comprehensive study on how to combine the real and generated data for future work.

\begin{table}[ht]
    \centering
    \small
    % \vspace{-0.15in}
    \caption{Test accuracy of ResNet18 trained on real data only, generated only, and a combination of real and generated data. }
    \label{tab:real_gen}
    \begin{tabular}{C{2.2cm}C{1.2cm}C{1.2cm}C{1.2cm}C{1.2cm}C{1.2cm}C{1.2cm}C{1.2cm}C{1.2cm}}
    \toprule
     & CIFAR-10 & CIFAR-100 & STL10 & Image-nette & Path-MNIST & Derma-MNIST & Blood-MNIST \\
    \midrule
    Real Only & 95.1 & 77.9 & 83.3 & 93.8 & 89.6 & \textbf{67.5} & \textbf{96.4} \\
    Generated Only & 94.6 & 74.4 & 89.0 & \textbf{95.4} & 82.1 & \textbf{67.5} & 93.7  \\
    Real + Generated & \textbf{95.2} & \textbf{78.0} & \textbf{90.0} & 95.0 & \textbf{92.1} & 66.3 & 95.7  \\
    \bottomrule
    \end{tabular}
    % \vspace{-0.2in}
\end{table}

\clearpage
\newpage
\subsection{FID, Precision, Recall, Density, and Coverage}
We assess the FID, precision, recall, density, and coverage of our generated data on STL10 using the implementation from \url{https://github.com/POSTECH-CVLab/PyTorch-StudioGAN}.

\paragraph{Interpolation Strength}
Table \ref{tab:alpha} demonstrates the variations in FID, precision, recall, density, and coverage with respect to the interpolation strength $\alpha$. As indicated, increasing the interpolation strength adversely affects FID, precision, and density, while improving recall. Coverage peaks at an interpolation of 0.1, suggesting a trade-off between generation quality and diversity.

\begin{table}[ht]
    \centering
    \caption{Image Generation Evaluation Metrics vs Interpolation Strength.}
    \label{tab:noise}
    \begin{tabular}{cccccc}
    \toprule
     alpha & FID & Precision & Recall & Density & Coverage\\
    \midrule
    0.00 & 17.930 & 0.894 & 0.644 & 0.734 & 0.753\\
    0.10 & 17.678 & 0.831 & 0.661 & 0.732 & 0.787\\
    0.20 & 26.177 & 0.635 & 0.751 & 0.584 & 0.739\\
    0.30 & 43.160 & 0.448 & 0.787 & 0.363 & 0.605\\
    0.40 & 62.773 & 0.328 & 0.805 & 0.245 & 0.500\\
    \bottomrule
    \end{tabular}
\end{table}

\paragraph{Gaussian Noise Strength}
Table \ref{tab:noise} demonstrates the variations in FID, precision, recall, density, and coverage as the additive noise value increases. The results indicate that higher noise levels adversely impact all metrics, signifying a decline in individual image quality. However, Figure 9 reveals that incorporating some noise can enhance model accuracy, as the overall information in the dataset may still increase despite the diminished quality of each image.

\begin{table}[ht]
    \centering
    \caption{Image Generation Evaluation Metrics vs Noise Value.}
    \label{tab:alpha}
    \begin{tabular}{cccccc}
    \toprule
     Noise Value & FID & Precision & Recall & Density & Coverage\\
    \midrule
    0.00 & 12.002 & 0.898 & 0.984 & 0.740 & 0.968\\
    0.10 & 13.210 & 0.865 & 0.979 & 0.698 & 0.947\\
    0.20 & 19.981 & 0.727 & 0.949 & 0.545 & 0.860\\
    0.30 & 38.839 & 0.476 & 0.893 & 0.294 & 0.614\\
    0.40 & 76.255 & 0.208 & 0.851 & 0.094 & 0.251\\
    \bottomrule
    \end{tabular}
\end{table}

\paragraph{Comparison against LECF}
Table \ref{tab:lecf} compares FID, precision, recall, density, and coverage between our method and LECF across various clip filter thresholds. Our approach outperforms LECF in all metrics, indicating that while choosing an optimal threshold improves the baseline LECF results, our method excels at generating high-quality, diverse images.

\begin{table}[ht]
    \centering
    \caption{Our method outperforms LECF in all metrics, suggesting that while selecting the optimal threshold enhances baseline LECF outcomes, our approach excels in generating higher quality and more diverse images.}
    \label{tab:lecf_app}
    \begin{tabular}{cccccc}
    \toprule
     Name & FID & Precision & Recall & Density & Coverage\\
    \midrule
    LECF (threshold=0.0) & 40.852 & 0.552 & 0.415 & 0.585 & 0.431\\
    LECF (threshold=0.1) & 40.858 &	0.552 &	0.431 &	0.591 &	0.432\\
    LECF (threshold=0.3) & 38.107 &	0.576 &	0.416 &	0.626 &	0.445\\
    LECF (threshold=0.5) & 37.061 &	0.589 &	0.413 &	0.641 &	0.449\\
    LECF (threshold=0.7) & 35.950 &	0.602 &	0.412 &	0.663 &	0.464\\
    LECF (threshold=0.9) & 34.522 &	0.631 &	0.416 &	0.708 &	0.477\\
    LECF (threshold=0.95) & 33.606 &	0.648 &	0.392 &	0.731 &	0.486\\
    LECF (threshold=0.97) & 33.224 &	0.664 &	0.381 &	0.756 &	0.490\\
    Diffusion Inversion (Ours) & 17.678 & 0.831 & 0.661 & 0.732 & 0.787\\
    \bottomrule
    \end{tabular}
\end{table}

\clearpage
\newpage
\subsection{Gaussian Noise}\label{app:gauss}
We investigate the influence of Gaussian noise on model performance by adjusting the noise strength and setting the latent interpolation strength $\alpha$ to $0$. Figure~\ref{fig:tsteps}c demonstrates the relationship between Gaussian noise strength and model test accuracy. Our findings indicate that the optimal performance is achieved when generating a dataset of equal size to the original without noise perturbation. However, when sampling additional data, it is advantageous to increase the noise strength accordingly, with a noise strength of $\lambda=0.2$ as a suitable starting point. 

\begin{figure}[ht]
\vskip 0.1in
\begin{center}
\centerline{\includegraphics[width=1\columnwidth]{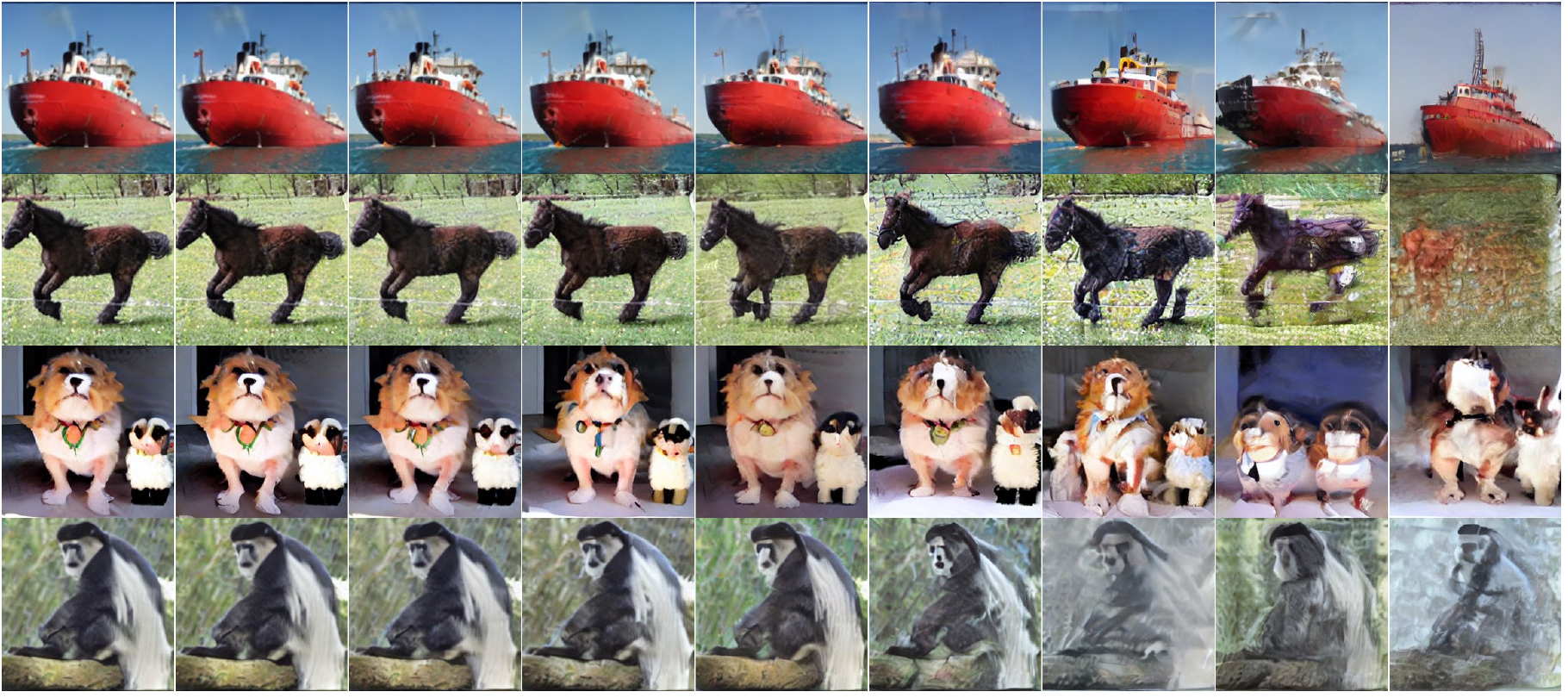}}
\caption{Generate image variants by perturbing the embedding vector using random Gaussian noise. Noise strength $\lambda$ from left to right: 0.0, 0.05, 0.10, 0.15, 0.20, 0.25, 0.30, 0.35, 0.40. The generated images of some embeddings are still meaningful under high strength. \vspace{-0.3in}}
\label{fig:ab_gaussian_vis}
\end{center}
% \vskip -0.1in
\end{figure}

Figure \ref{fig:ab_gaussian_vis} presents the generated images at varying noise levels, showing minimal differences between perturbed and original images when the noise level is below $\lambda=0.2$. Nonetheless, significant variations are observed at higher noise levels, such as the ship image remaining discernible at $\lambda=0.4$, while the horse becomes indistinguishable. Ideally, we may want to employ distinct Gaussian noise strengths for each image rather than using a single fixed value for all.

\clearpage
\newpage
\subsection{Additional Visualization}
\begin{figure*}[ht]
\begin{center}
\begin{subfigure}[b]{0.49\textwidth}
\centering
\includegraphics[width=1.0\linewidth]{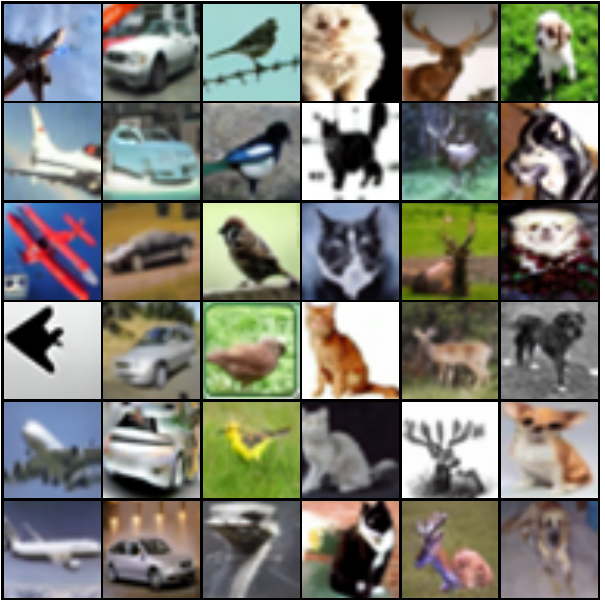}
\caption{CIFAR10}
\end{subfigure}
\hfill
\begin{subfigure}[b]{0.49\textwidth}
\centering
\includegraphics[width=1.0\linewidth]{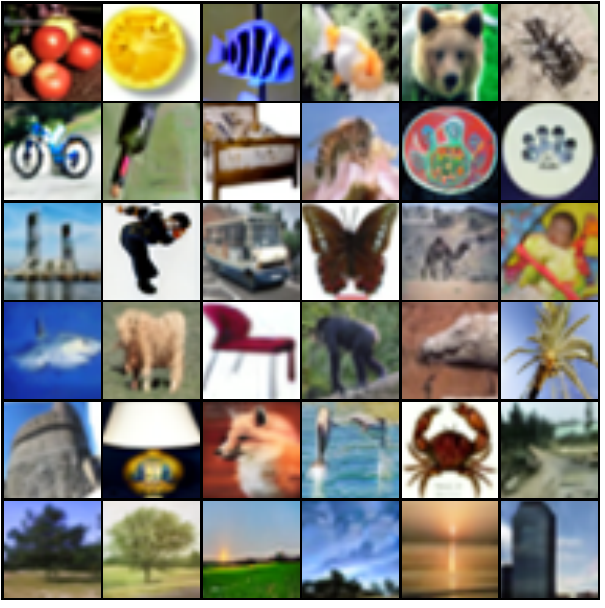}
\caption{CIFAR100}
\end{subfigure}
\begin{subfigure}[b]{0.49\textwidth}
\centering
\includegraphics[width=1.0\linewidth]{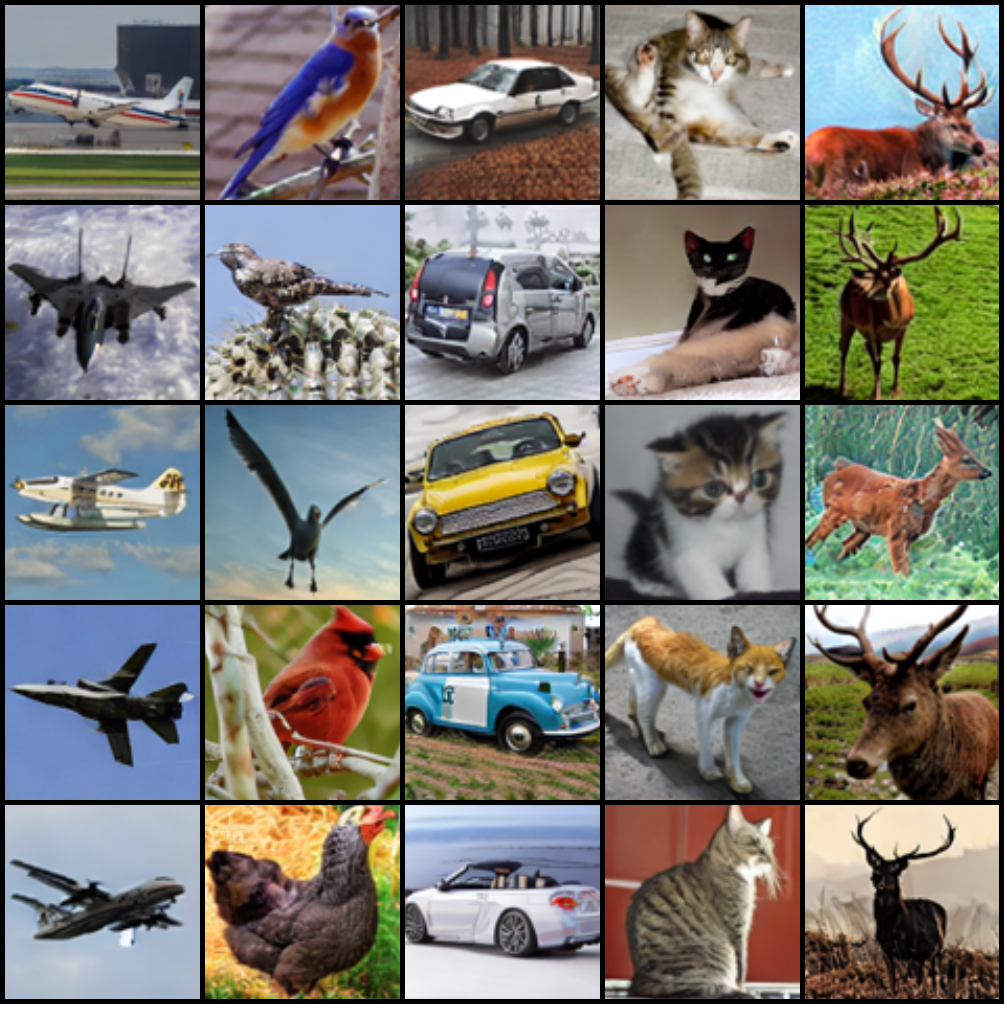}
\caption{STL10}
\end{subfigure}
\hfill
\begin{subfigure}[b]{0.49\textwidth}
\centering
\includegraphics[width=1.0\linewidth]{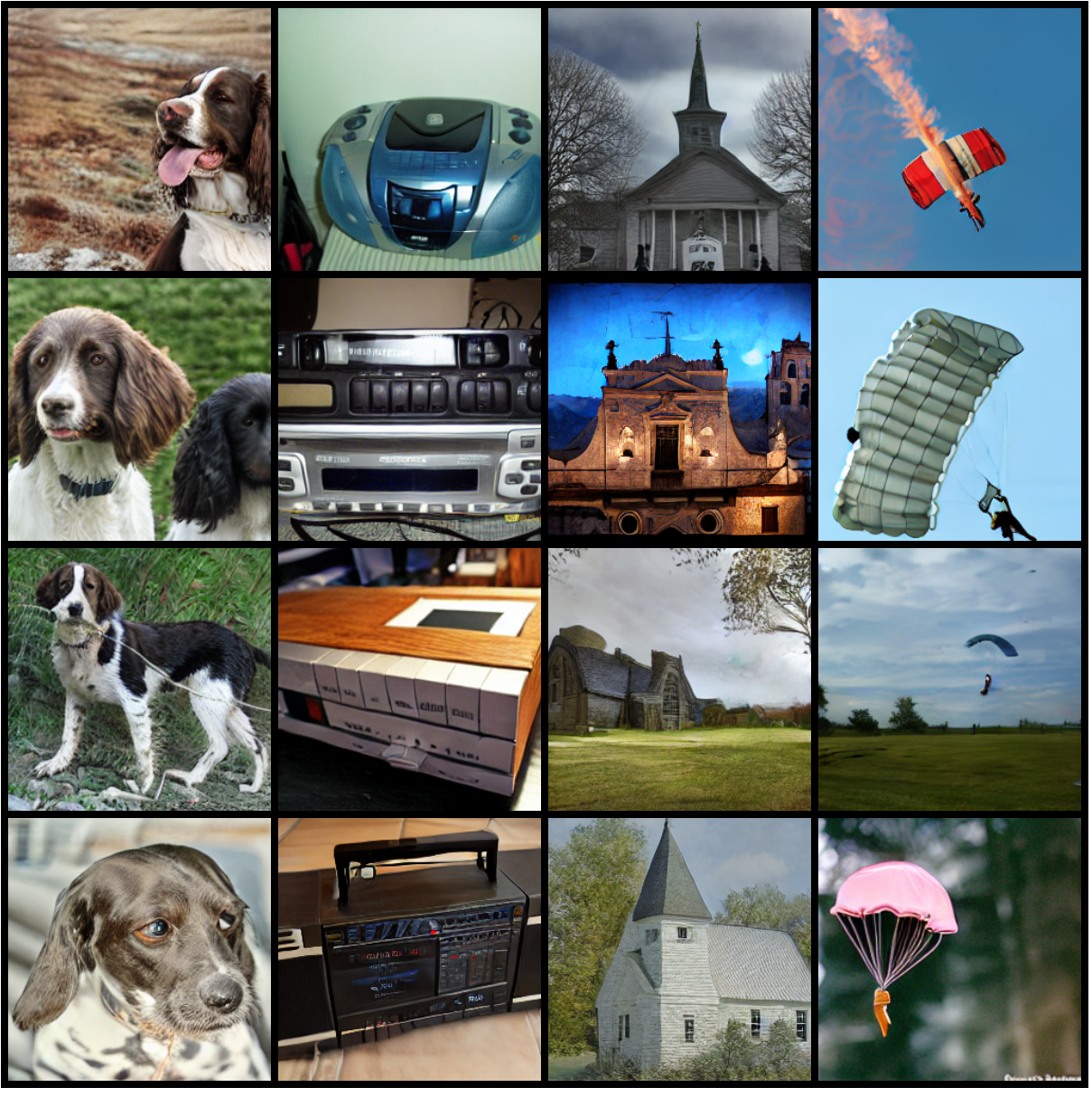}
\caption{ImageNette}
\end{subfigure}
\end{center}
\caption{\footnotesize{Synthetic images produced by our method: exhibiting diversity, realism, and comprehensive representation of the original dataset, effectively serving as a suitable substitute.}}
\label{fig:syn_data_vis2}
\end{figure*}

\begin{figure*}[ht]
\begin{center}
\begin{subfigure}[b]{0.49\textwidth}
\centering
\includegraphics[width=1.0\linewidth]{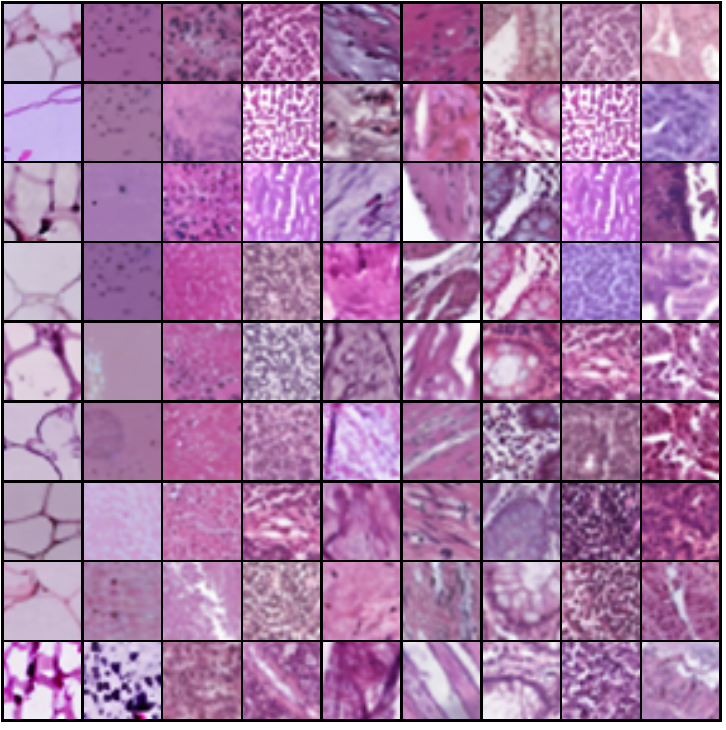}
\caption{PathMNIST}
\end{subfigure}
\hfill
\begin{subfigure}[b]{0.49\textwidth}
\centering
\includegraphics[width=1.0\linewidth]{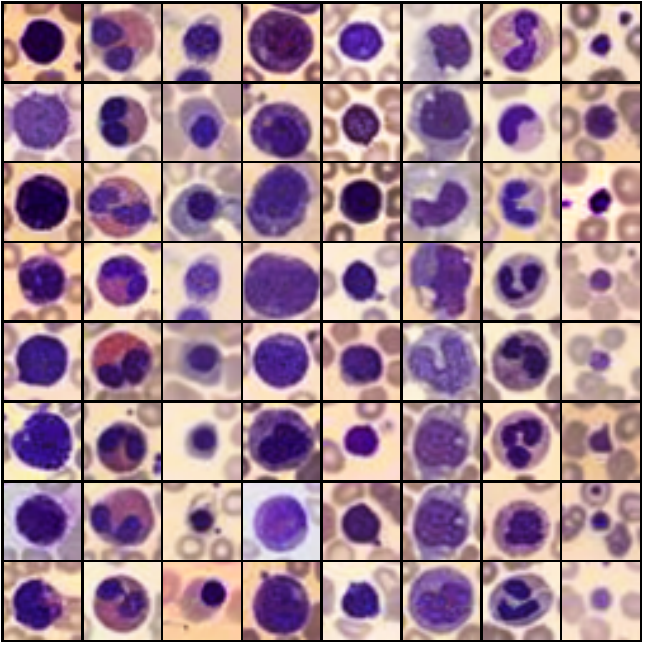}
\caption{BloodMNIST}
\end{subfigure}
\begin{subfigure}[b]{0.49\textwidth}
\centering
\includegraphics[width=1.0\linewidth]{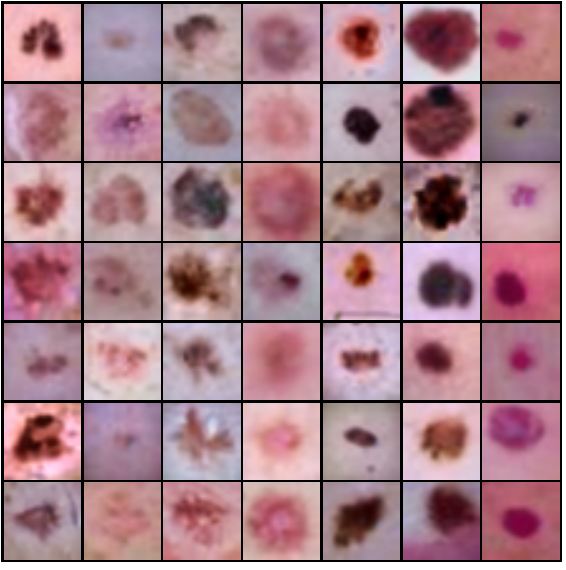}
\caption{DermaMNIST}
\end{subfigure}
\hfill
\begin{subfigure}[b]{0.49\textwidth}
\centering
\includegraphics[width=1.0\linewidth]{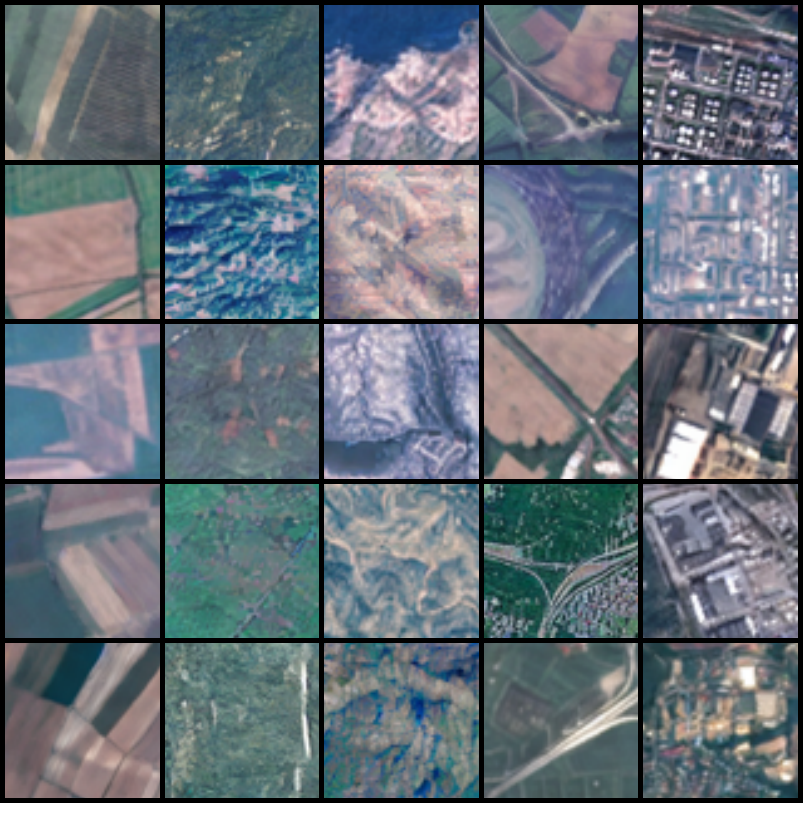}
\caption{EuroSAT}
\end{subfigure}
\end{center}
\caption{\footnotesize{Synthetic images produced by our method: exhibiting diversity, realism, and comprehensive representation of the original dataset, effectively serving as a suitable substitute.}}
\label{fig:syn_data_vis3}
\end{figure*}
%%%%%%%%%%%%%%%%%%%%%%%%%%%%%%%%%%%%%%%%%%%%%%%%%%%%%%%%%%%%

\end{document}